\documentclass[sigconf]{acmart}

\makeatletter
\renewcommand\@formatdoi[1]{\ignorespaces}
\makeatother

\usepackage{algorithmicx,algorithm}
\usepackage[noend]{algpseudocode}
\usepackage{amsmath}
\usepackage{graphicx}
\usepackage{subfigure}
\usepackage{caption}
\usepackage{graphicx}
\usepackage{multirow}

\AtBeginDocument{%
  \providecommand\BibTeX{{%
    \normalfont B\kern-0.5em{\scshape i\kern-0.25em b}\kern-0.8em\TeX}}}

\setcopyright{acmcopyright}
\copyrightyear{2022}
\acmYear{2022}
\setcopyright{acmlicensed}
\acmConference[KDD '22]{Proceedings of the 28th ACM SIGKDD Conference on Knowledge Discovery and Data Mining}{August 14--18, 2022}{Washington, DC, USA}

\acmBooktitle{Proceedings of the 28th ACM SIGKDD Conference on Knowledge Discovery and Data Mining (KDD '22), August 14--18, 2022, Washington, DC, USA}

\algrenewcommand\algorithmicrequire{\textbf{Input:}}
\algrenewcommand\algorithmicensure{\textbf{Output:}}

\begin{document}
\fancyhead{}
\title{Collaborative Intelligence Orchestration: Inconsistency-Based Fusion of Semi-Supervised Learning and Active Learning}

\author{Jiannan Guo}
\affiliation{%
\institution{Zhejiang University}
\city{}
\country{}
}
\email{jiannan@zju.edu.cn}

\author{Yangyang Kang}
\affiliation{%
\institution{Alibaba Group}
\city{}
\country{}
}
\email{yangyang.kangyy@alibaba-inc.com}

\author{Yu Duan}
\affiliation{%
\institution{Alibaba Group}
\city{}
\country{}
}
\email{derrick.dy@alibaba-inc.com}

\author{Xiaozhong Liu}
\affiliation{%
\institution{Indiana University Bloomington}
\city{}
\country{}
}
\email{liu237@indiana.edu}

\author{Siliang Tang}
\authornote{Corresponding Author.}
\affiliation{%
\institution{Zhejiang University}
\city{}
\country{}
}
\email{siliang@zju.edu.cn}

\author{Wenqiao Zhang}
\affiliation{%
\institution{Zhejiang University}
\city{}
\country{}
}
\email{wenqiaozhang@zju.edu.cn}

\author{Kun Kuang}
\affiliation{%
\institution{Zhejiang University}
\city{}
\country{}
}
\email{kunkuang@zju.edu.cn}

\author{Changlong Sun}
\affiliation{%
\institution{Alibaba Group}
\city{}
\country{}
}
\email{changlong.scl@taobao.com}

\author{Fei Wu}
\affiliation{%
\institution{Zhejiang University}
\city{}
\country{}
}
\email{wufei@zju.edu.cn}

\begin{abstract}
While annotating decent amounts of data to satisfy sophisticated learning models can be cost-prohibitive for many real-world applications. Active learning (AL) and semi-supervised learning (SSL) are two effective, but often isolated, means to alleviate the data-hungry problem. Some recent studies explored the potential of combining AL and SSL to better probe the unlabeled data. However, almost all these contemporary SSL-AL works use a simple combination strategy, ignoring SSL and AL's inherent relation. Further, other methods suffer from high computational costs when dealing with large-scale, high-dimensional datasets. Motivated by the industry practice of labeling data, we propose an innovative \textbf{I}nconsistency-based virtual a\textbf{D}v\textbf{E}rsarial \textbf{A}ctive \textbf{L}earning (IDEAL) algorithm to further investigate SSL-AL's potential superiority and achieve mutual enhancement of AL and SSL, \emph{i.e.,} SSL propagates label information to unlabeled samples and provides smoothed embeddings for AL, while AL excludes samples with inconsistent predictions and considerable uncertainty for SSL. We estimate unlabeled samples' inconsistency by augmentation strategies of different granularities, including fine-grained continuous perturbation exploration and coarse-grained data transformations. Extensive experiments, in both text and image domains, validate the effectiveness of the proposed algorithm, comparing it against state-of-the-art baselines. Two real-world case studies visualize the practical industrial value of applying and deploying the proposed data sampling algorithm.
\end{abstract}

\ccsdesc[500]{Computing methodologies~Active learning settings}
\ccsdesc[500]{Computing methodologies~Semi-supervised learning settings}

\keywords{active learning, semi-supervised learning, adversarial learning}


\maketitle

\section{Introduction}

The utilization of large training datasets is an essential ingredient to endorse the success of deep learning~\cite{li2021adaptive,li2022compositional,zhang2021magic,zhang2021consensus} and its applications (\emph{e.g.,} image classification~\cite{krizhevsky2017imagenet}, text classification~\cite{chen2017recurrent,de2020victor}).
However, in real-world systems, the acquisition of decent labeled data is cost-prohibitive and challenging (\emph{e.g.,} legal text annotation~\cite{xu2020distinguish} can be expensive because of domain expert scarcity). Therefore, initiating deep learning with sparse~\cite{li2022end} and ever-growing training data, with/without human involvement, can be a vital task to explore.

\begin{figure*}[htbp]
\centering
\subfigure[]{
\begin{minipage}[t]{0.32\linewidth}
\centering
\includegraphics[width=2.3in]{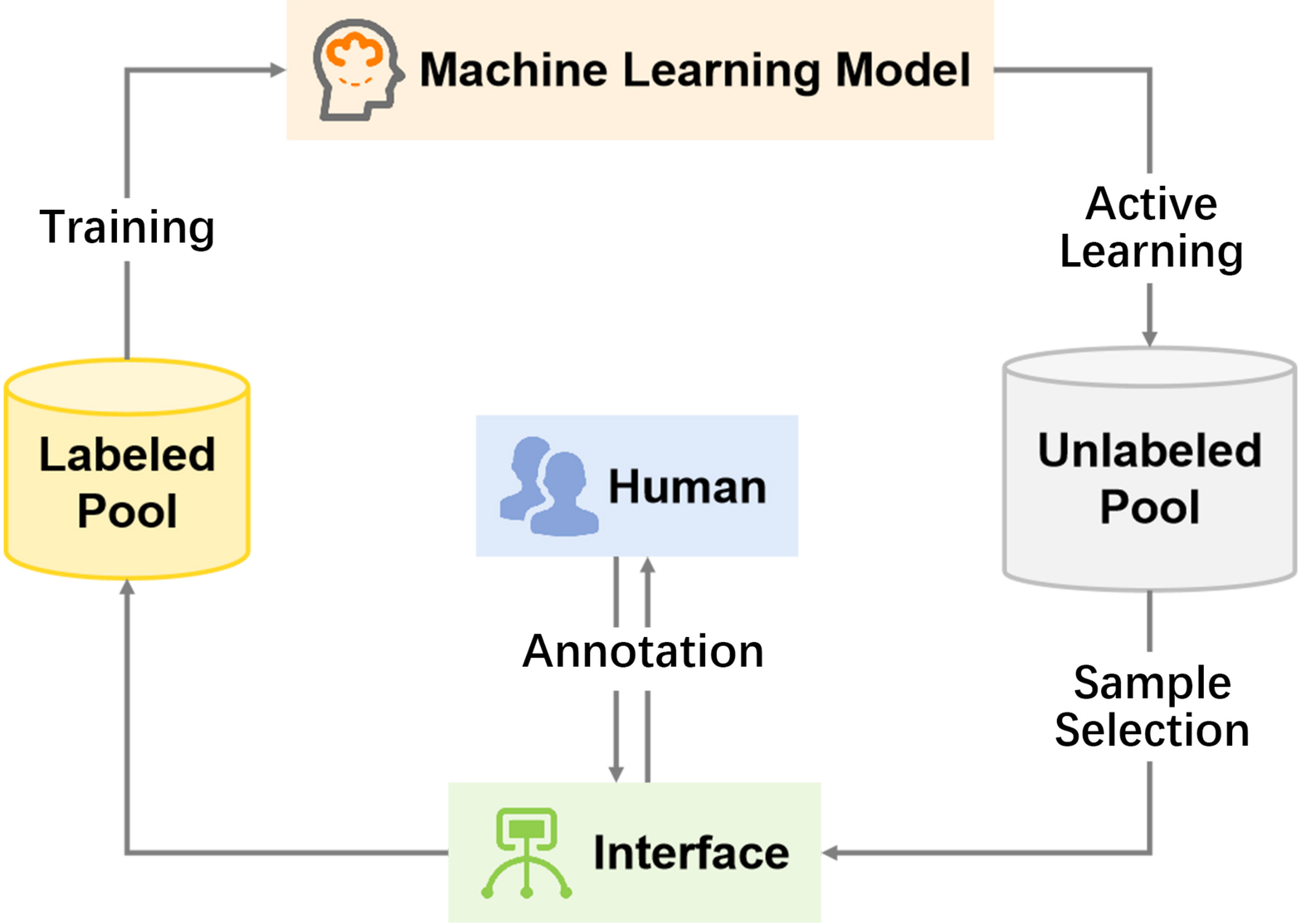}
\label{p1}
\end{minipage}%
}%
\subfigure[]{
\begin{minipage}[t]{0.32\linewidth}
\centering
\includegraphics[width=2in]{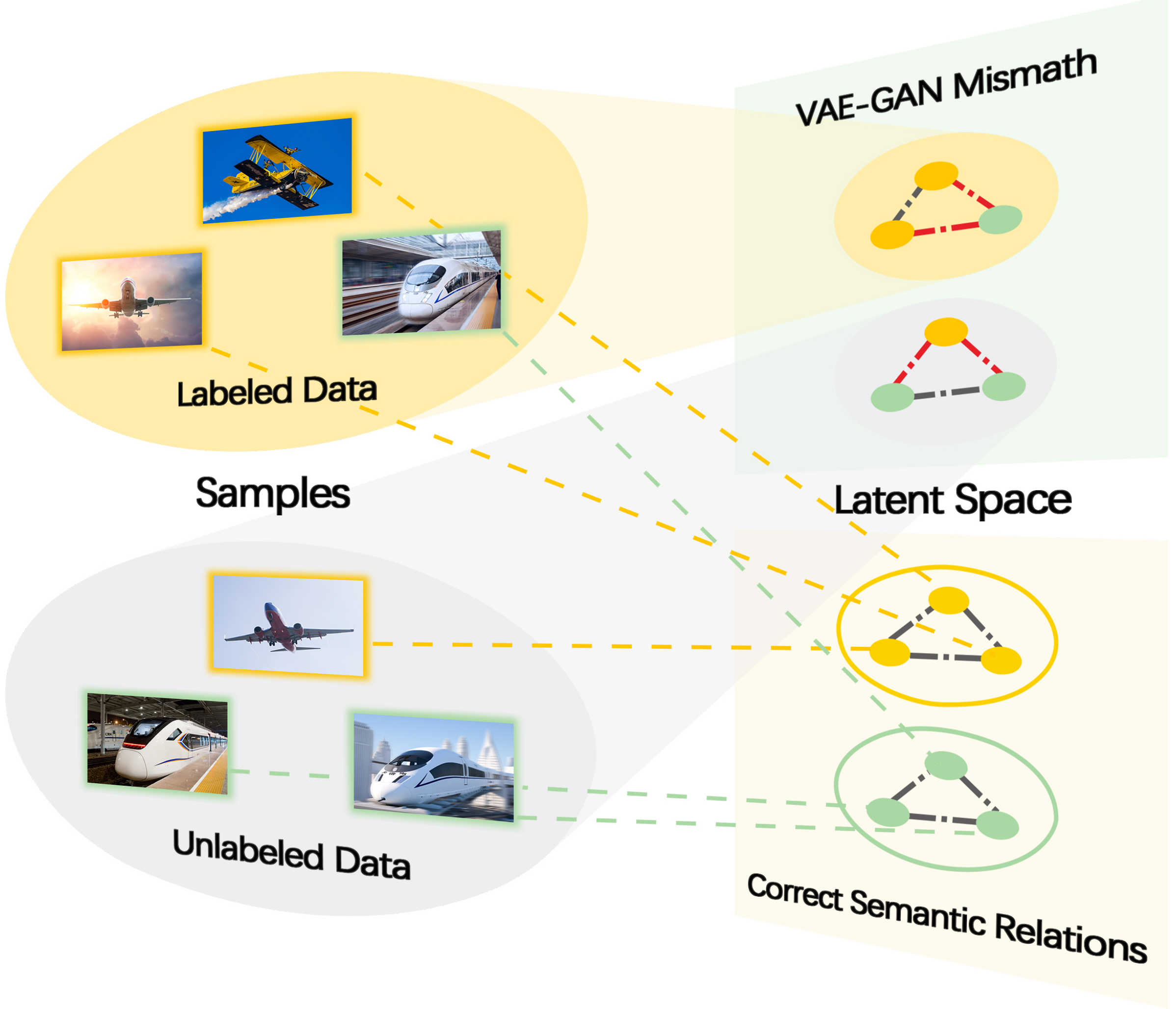}
\label{p2}
\end{minipage}%
}%
\subfigure[]{
\begin{minipage}[t]{0.32\linewidth}
\centering
\includegraphics[width=2in]{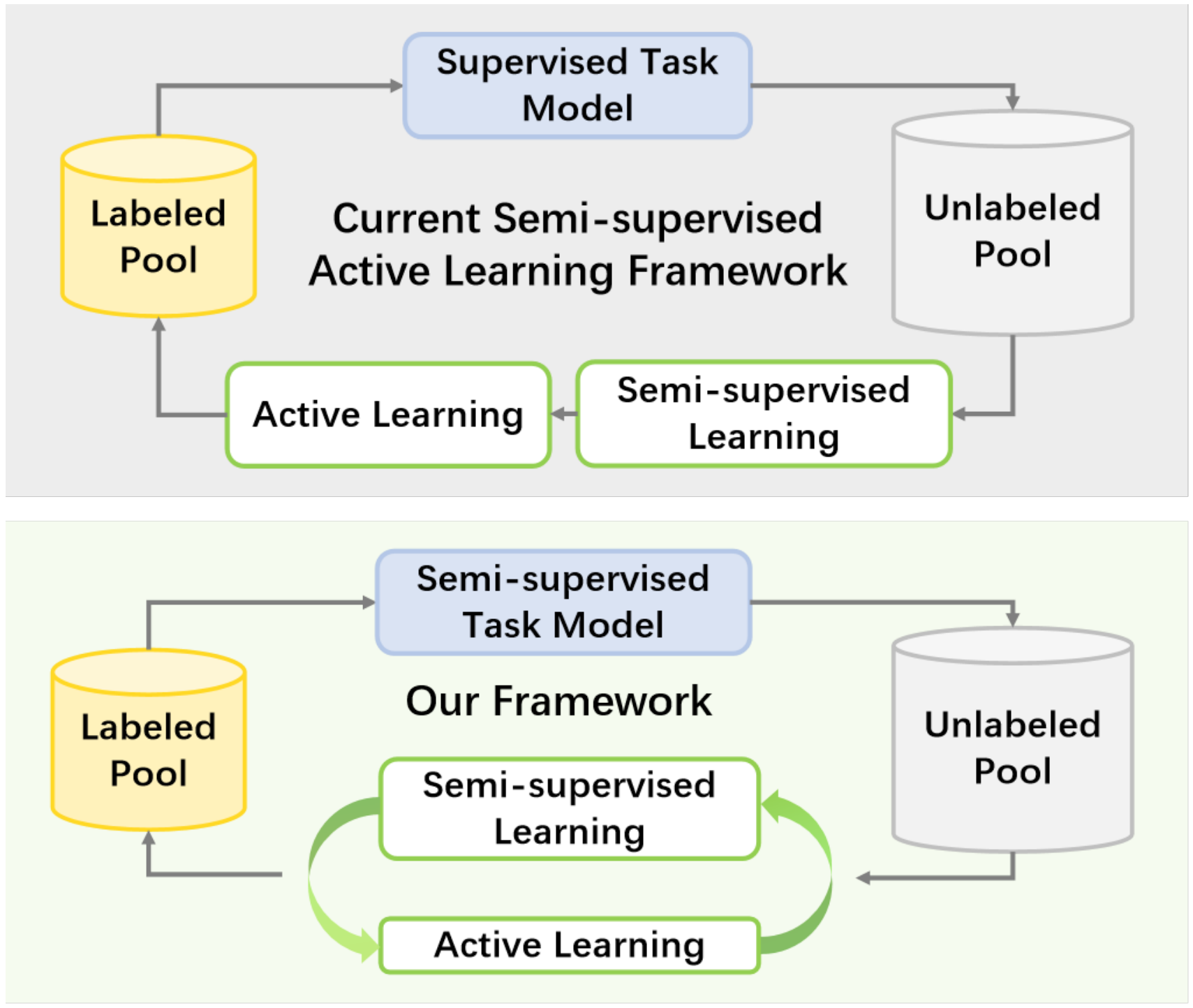}
\label{p3}
\end{minipage}
}%
\vspace{-0.35cm}
\centering
\caption{(a) Active learning illustration. (b) Since samples' labeling states and class labels are uncorrelated, these methods tend to project the representation of samples with different class labels (\emph{e.g.,} train and plane) to the same class (\emph{i.e.,} labeled or unlabeled), and vice versa. (c) Comparison of current SSL-AL methods and our approach.}
\vspace{-0.3cm}
\end{figure*}

To address this problem, prior studies embraced two directions:
1) \emph{active learning, (AL)}~\cite{pmlr-v70-gal17a,mayer2020adversarial,yoo2019learning} aims to select the most informative samples for human annotation with the least labeling cost (Figure~\ref{p1}); 2)\emph{semi-supervised learning, (SSL)}~\cite{berthelot2019mixmatch,kipf2016semi,chen2020mixtext} also faces such challenges with labeled data sparseness and unlabeled data abundance.
Along this line of research, the potential of combining SSL and AL (\textit{i.e.,} SSL-AL) can be nontrivial to further improve model performance. However, recent SSL-AL investigations are limited by the following two drawbacks: 1). Most current SSL-AL methods~\cite{zhang2020state,sinha2019variational,Kim_2021_CVPR} with VAE-GAN structure suffer from the mismatch problem (Figure~\ref{p2}), as learned representations of samples does not facilitate (or even do harm to) the classification. 2). Current methods~\cite{zhang2020state,sinha2019variational,Kim_2021_CVPR} employ SSL and AL as two isolated modules, while great potentials can be unleashed by enabling the sophisticated interaction between SSL and AL (Figure~\ref{p3}). 

Indeed, AL and SSL can be complementary when they share the same goal (\textit{i.e.}, utilizes unlabeled samples), and they can work collaboratively to achieve mutual enhancement:
1) (SSL$\rightarrow$AL), SSL can provide enhanced embeddings to assist AL to exploit unlabeled samples' underlying distribution and evaluate unlabeled samples' attributes;
2) (AL$\rightarrow$SSL), AL can exclude some complex examples from the unlabeled pool and help SSL reduce the uncertainty of model's prediction.
Utilizing the potential relatedness among various learning methods can be regarded as a form of inductive transfer, \textit{e.g.}, \emph{inductive bias}~\cite{baxter2000model}, to equip the combined learning model with the correct hypothesis and performance improvement. SSL-AL reciprocal optimization is a novel but critical problem for human-in-the-loop AI.


In this work, we utilize `sample inconsistency' as the bridge to enable the interaction between SSL and AL, which is inspired by the fact that the sample inconsistency is an effective means for both SSL~\cite{berthelot2019mixmatch} and AL~\cite{gao2020consistency}. The robustness against the sample inconsistency (\emph{e.g.,} the translation invariance of CNNs) can provide essential information to enhance models' generalization ability. Furthermore, the prior study~\cite{miyato2018virtual} has shown that not only human-perceivable but also human-unperceivable sample variance can lead to a significant impact on models' prediction.  

To address these challenges, we propose a novel \textbf{I}nconsistency-based virtual a\textbf{D}v\textbf{E}rsarial \textbf{A}ctive \textbf{L}earning (IDEAL) framework to enhance human-in-the-loop AI by leveraging effective SSL-AL fusion. IDEAL carries three modules: \textbf{the SSL label propagator} propagates the label information from the sparse training data to the unlabeled samples, which is a mixture of coarse-grained (human-perceivable) and fine-grained (human-unperceivable) augmentations. It yields the task model with high consistency and low entropy. Then, \textbf{the virtual inconsistency ranker} ranks all the unlabeled samples in terms of their coarse-grained and fine-grained inconsistency and selects top samples (\emph{i.e.,} poor consistency) as the rough annotation candidates. Finally, \textbf{the density-aware uncertainty re-ranker} further filters the selected samples (\emph{i.e.,} high entropy), and provides the final samples for human annotation.

To validate the performance of IDEAL, we provide a comprehensive evaluation on four benchmark datasets and two real-world datasets.
Experiments demonstrate that IDEAL outperforms state-of-the-art approaches and makes noticeable economic benefits.

Our major contribution can be summarized as follows:
\begin{itemize}
	\item
	We propose a novel \textbf{I}nconsistency-based virtual a\textbf{D}v\textbf{E}rsarial \textbf{A}ctive \textbf{L}earning (IDEAL) framework where SSL and AL form a closed-loop structure for reciprocal optimization and enhancement.
	\item
	We develop novel coarse-grained and fine-grained data augmentation strategies and couple them to both SSL and AL. Specially, we leverage the virtual adversarial technique to explore the continuous local distribution of unlabeled samples and further measure the inconsistency of samples.
	\item We evaluate IDEAL over diverse text and image datasets. The results validate IDEAL is superior than previous approaches and it is cost effective in various industrial applications.  
\end{itemize}

\section{Related Work}

In practice, it is easy to acquire abundant unlabeled samples. Thus pool-based 	AL~\cite{zhang2020state,yoo2019learning,sener2017active,
sinha2019variational} is more popular than the other two scenarios: steam-based ~\cite{dagan1995committee} and membership query synthesis ~\cite{tran2019bayesian}.

\noindent $\bullet$ \textbf{Pool-based active learning.}\quad
Uncertainty-based sampling and distribution-based sampling are common methods in the pool-based scenario. Our
method considers both uncertainty and distribution. For uncertainty-based methods, the framework~\cite{kapoor2007active} uses Gaussian
processes to evaluate uncertainty, while ~\cite{Ebrahimi2020Uncertainty-guided} uses a Bayesian network. MC dropout~\cite{pmlr-v70-gal17a}
is used as an approximation of the Bayesian network. However, these methods have a high computational cost and can not handle large-scale datasets. In the era of deep-learning-based AL,~\cite{zhang2017active} and~\cite{yoo2019learning} adopt similar metrics to choose informative samples.~\cite{zhang2017active} selects samples leading to the greatest change to the model's gradient during the training process, while~\cite{yoo2019learning} selects samples with the biggest training loss. However, the task model's loss and gradient information are unstable in the early training stage, and they may influence the quality of the final selected samples.

The distribution-based method~\cite{sener2017active} selects samples of a subset whose feature distribution covers the entire feature space as much as possible. However, subset selection (NP-hard problem) becomes computationally infeasible for large-scale datasets or high-dimension input data.
Thus, the algorithms of selecting subsets will suffer from inefficient computing.  

\noindent $\bullet$ \textbf{Semi-supervised active learning.}\quad
Recently, ~\cite{sinha2019variational, Kim_2021_CVPR,zhang2020state} leverage VAE-GAN structure to learn the representation of both labeled and unlabeled samples in latent space. However, these
methods misuse the relation between labeling states and class labels. Consequently, the learning process may harm the semantic distribution of samples in latent space. Annotation information (labeling states) and the feature distribution (class labels) are orthogonal, while the feature distribution is highly correlated with semantic information of different classes. 
Compared to these VAE-GAN based methods, our method considers inconsistency of different granularity to  combine AL and SSL. 
The works ~\cite{song2019combining, gao2020consistency} combine AL and SSL based on prediction consistency given a set of data augmentations. However, these methods only use a limited number of ways of human-perceivable data augmentation to estimate inconsistency. In contrast, our method further leverages human-unperceivable inconsistency to obtain more abundant information of unlabeled samples for model estimation and optimization. Besides, we also consider the feature distribution of samples. 

Our method is related to the latest work~\cite{guo2021semi} in terms of hierarchical sample selection. Sadly,~\cite{guo2021semi} suffers from high computational and spatial costs caused by its critical step (builds a KNN graph in every selection cycle). Consequently, deploying~\cite{guo2021semi} in real-world scenarios to handle large-scale datasets faces enormous challenges. In addition,~\cite{guo2021semi} does not integrate adversarial perturbation generation into the training process. Thus, the selection criterion (adversarial perturbation) can not adapt dynamically to the model's optimization. Our method is free of the above two deficiencies. Firstly, we replace graph SSL with the Mixup technique to handle large-scale datasets. Then we couple selection criterion (fine-grained inconsistency) to model optimization for optimal perturbation generation. 

Unlike our work, all of these methods are specially designed for image classification tasks, which can not be directly transferred to tasks of other media types (\textit{e.g.,} text type).

\section{Method}

In this section, we formulate the proposed \textbf{I}nconsistency-based virtual a\textbf{D}v\textbf{E}rsarial \textbf{A}ctive \textbf{L}earning (IDEAL) algorithm.
We first provide a brief overview of our whole AL framework, and then introduce three main components of IDEAL: a SSL label propagator~(section ~\ref{sec:slp}), a virtual inconsistency ranker~(section~\ref{sec:vad}) and a density-aware uncertainty re-ranker~(section~\ref{sec:bl}).

\begin{figure*}[t]
\includegraphics[width=0.95\textwidth]{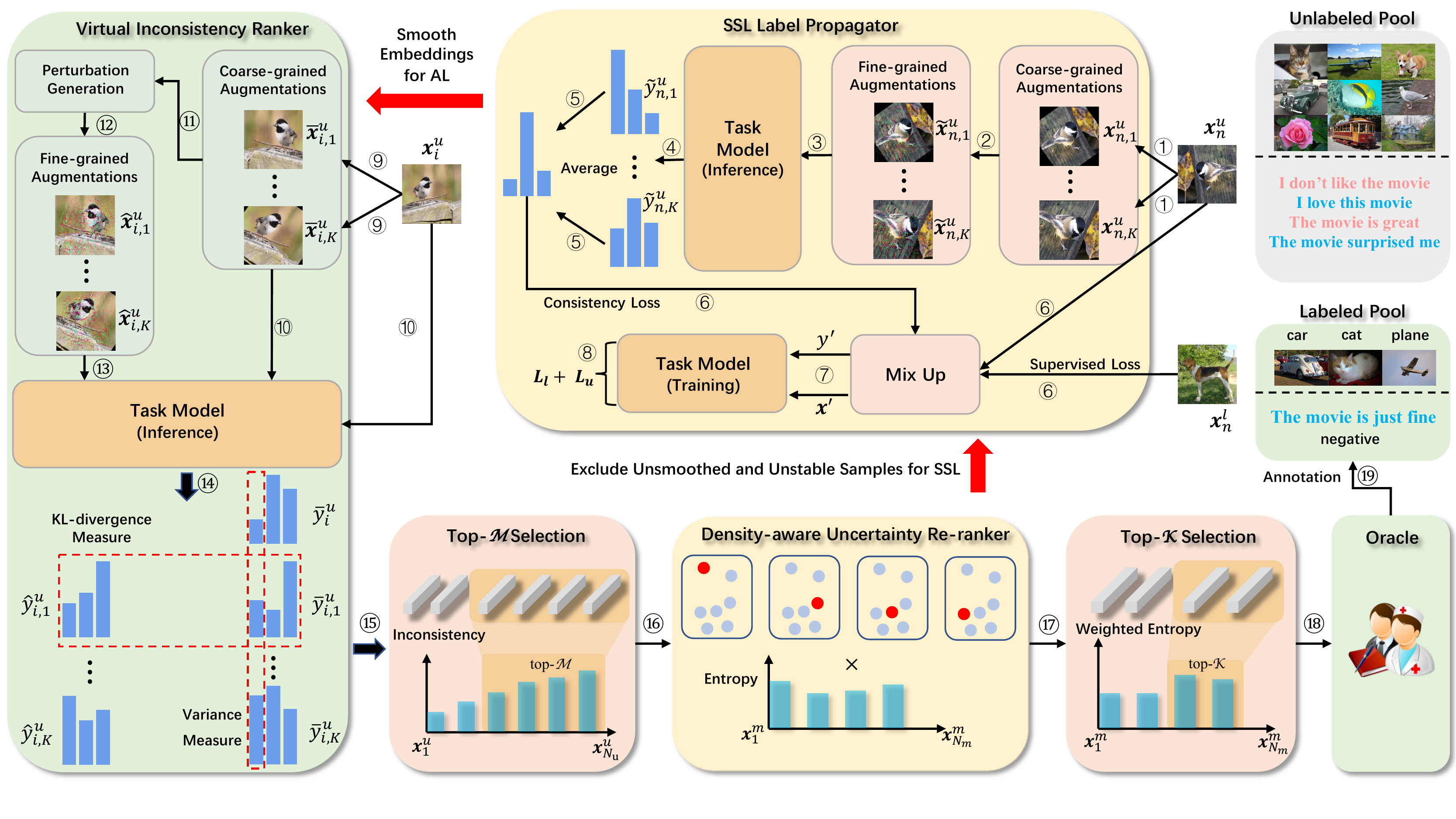}
\vspace{-0.3cm}
\centering\caption{An overview of our proposed AL framework IDEAL which consists of three modules: (a) \emph{SSL Label Propagator} propagates label information from labeled samples to unlabeled samples and smooths local distribution for AL;
 (b) \emph{Virtual Inconsistency Ranker} estimates inconsistency by calculating coarse-grained and fine-grained inconsistency respectively;
(c) \emph{Density-aware Uncertainty Re-ranker} further screens the samples with high density-aware uncertainty. }
\vspace{-0.3cm}
\label{fig:arc}
\end{figure*}

\subsection{Overview}\label{sec:ovr}
In this section, we formally describe the pool-based AL loop with our proposed IDEAL demonstrated in Figure~\ref{fig:arc}.
We define the labeled pool as $\mathcal{D}^l = \{(\textbf{x}^l_{1}, y^l_{1}), \dots, (\textbf{x}^l_{N_l}, y^l_{N_l})\}$ and the unlabeled pool as $\mathcal{D}^u= \{\textbf{x}^u_1, \dots, \textbf{x}^u_{N_u}\}$ ($N_l$ and $N_u$ are the numbers of labeled samples and unlabeled samples respectively).
(1) For samples in the pool, IDEAL first feeds them into the SSL label propagator, which propagates the label information from labeled samples to unlabeled samples by mixing up samples. (2) Based on the inconsistency signals provided by the SSL label propagator, our virtual inconsistency ranker calculates the total inconsistency (coarse-grained and fine-grained inconsistency) for each unlabeled sample in the pool $\mathcal{D}^u$, and selects the top-$\mathcal{M}$ samples with the largest inconsistency as initial annotation candidates. 
(3) The density-aware uncertainty re-ranker further selects top-$\mathcal{K}$ candidates with the largest density-aware uncertainty from initial top-$\mathcal{M}$ candidates for human annotation, finally excluding unsmoothed and unstable samples for SSL to perform better label propagation. As a consequence, the sizes of the labeled pool and the unlabeled pool will be updated to $N_l+\mathcal{K}$ and $N_u - \mathcal{K}$ respectively. The loop will be repeated until the annotation budget is run out.

\subsection{SSL Label Propagator}
\label{sec:slp}
We leverage the SSL label propagator to propagate the label information from labeled samples to unlabeled samples by smoothing local inconsistency distribution of unlabeled samples, and obtain enhanced embeddings for following AL.

Formally speaking, given unlabeled dataset $\mathcal{D}^u= \{\textbf{x}^u_1, \dots, \textbf{x}^u_{N_u}\}$ and the labeled dataset $\mathcal{D}^l =\{(\textbf{x}^l_{1}, y^l_{1}), \dots, (\textbf{x}^l_{N_l}, y^l_{N_l})\}$, we firstly generate $K$ augmented unlabeled samples $\textbf{X}^u_n=\{\textbf{x}^u_{n,1}, \dots, \textbf{x}^u_{n,K}\}$ for each unlabeled sample $\textbf{x}^u_n\in \mathcal{D}^u$ through coarse-grained augmentations. Specifically, the coarse-grained augmentation is a family of data augmentation functions $T$ that s.t. $\textbf{x}_{aug} = T(\textbf{x})$ and $max(\textbf{x}_{aug} - \textbf{x}) > \delta$ (where $\delta$ is a threshold discriminating human-perceivable and human-unperceivable difference). The corresponding inconsistency estimation is called to be coarse-grained, and vice versa. 
Moreover, we adopt sentence-level transformations, rewriting sentences with similar meanings (back translation) as the coarse-grained augmentation strategy for text data. We use image-level transformations (\textit{e.g.}, rotation and clipping of the whole image) to obtain coarse-grained augmented image samples.

In addition to the above discrete coarse-grained data transformations, we introduce pixel-level or embedding-level fine-grained augmentations, adding local perturbation to inputs’ representations in feature space (described in section~\ref{sec:vad}), to force the model to explore continuous local distribution.

After applying these augmentation strategies of two different granularity, we can obtain the final augmentation samples $\Tilde{\textbf{X}}^u_n=\{\Tilde{\textbf{x}}^u_{n,1}, \dots, \Tilde{\textbf{x}}^u_{n,K}\}$ for samples in 
$\textbf{X}^u_n$.


Then, we feed each unlabeled sample $\textbf{x}^u_n$ and its augmented samples $\{\Tilde{\textbf{x}}^u_{n,1}, \dots, \Tilde{\textbf{x}}^u_{n,K}\}$ into the task model to
get predicted labels $\Tilde{y}_n^u$ and $\{\Tilde{y}^u_{n,1}, \dots, \Tilde{y}^u_{n,K}\}$. We generate shared guessed label for augmentation samples by averaging all of their predicted labels:

\begin{equation}
\Tilde{y}^a_n =  \frac{1}{w_u+\sum_kw_k}(w_u\Tilde{y}_n^u+\sum_kw_k\Tilde{y}^u_{n,k}).
\end{equation}
where weights $w_u$ and $w_k$ mean the contribution ratios of different augmentation samples to the final guessed label.
We perform a weighted average operation to guarantee the task model to output consistent predictions for different augmentation samples. We set weights parameters for original samples and their augmented samples based on translation qualities of different languages in text classification. For images, we set all weights parameters to 1.

%

Next, we obtain the labeled set $\mathcal{D}^l$, the unlabeled set $\mathcal{D}^u$ and the augmented set $\mathcal{D}^a=\{(\Tilde{\textbf{X}}^u_1, \Tilde{y}^a_1), \dots, (\Tilde{\textbf{X}}^u_{N_u}, \Tilde{y}^a_{N_u})\}$ ( $\mathcal{D}^u$ and $\mathcal{D}^a$ share the same labels). We can obtain training samples by randomly mixing two batches of samples' hidden states from these three sets as well as their respective labels (or guessed labels) together.

We can obtain three types of mixed samples: (1) mixing two batches of labeled samples; (2) mixing a batch of labeled and a batch of unlabeled samples; (3) and mixing two batches of unlabeled samples. This can be defined as:

\begin{equation}
\textbf{x}^{'} =  \lambda \textbf{h}_1 + (1-\lambda)\textbf{h}_2.
\end{equation}

\begin{equation}
y^{'} =  \lambda y_1 + (1-\lambda)y_2.
\end{equation}
where $\textbf{h}_1, \textbf{h}_2$ are the hidden states corresponding to the samples $\textbf{x}_1$ and $\textbf{x}_2$. In detail, for text tasks, we select the hidden states of the middle layer of the task model for mixing up. For image tasks, we mix up the raw input features directly. Besides, the weight factor $\lambda$ is sampled from \textit{Beta Distribution}, and it can be further defined as:

\begin{equation}
    \lambda^{'} = Beta(\alpha,\alpha).
\end{equation}

\begin{equation}
\lambda = max(\lambda^{'},1-\lambda^{'}).
\end{equation}
where $\alpha$ is the hyper-parameter of the $Beta$ distribution $\lambda^{'}$. After the mixing process, the mixed feature $\textbf{x}^{'}$ aggregates both labeled samples' and unlabeled samples' features. Similarly, the mixed label $y^{'}$ aggregates supervised label information from labeled samples and distribution information of unlabeled samples.







Finally, we define the loss function as $L_l + L_u$: cross-entropy loss term $L_l$ is minimized through supervised label information. Consistency loss term $L_u$ (\textit{e.g.}, KL-divergence or L2 norm) is applied to constrain different augmented samples to have the same label with their original unlabeled samples.

Trained with mixed samples, the label propagator will propagate rich label information from labeled samples to unlabeled samples. The propagator smoothes the local distribution of unlabeled samples and makes the task model insensitive to both coarse-grained data transformation and fine-grained perturbation, providing enhanced embeddings for AL. 
Mixup mines and optimizes local and global samples distribution for SSL and AL. Apart from the local distribution mining based on the smoothing of point-wise augmentation, Mixup explores convex combinations of pairs of labeled, unlabeled, and augmented samples to mine the global distribution. Since enumerating all the convex combinations of pairs of samples to precisely model the whole picture is intractable, we utilize the random sampling method to estimate the distribution and optimize the corresponding objectives to learn the desired distribution. Transformed/untransformed samples with similar classes are close.


\subsection{Virtual Inconsistency Ranker}\label{sec:vad}

Above SSL label propagator can smooth the local distribution around unlabeled samples, which means that the task model's predictions for unlabeled samples tend to be consistent before and after perturbation.
Our goal is to select the samples with poor consistency for further annotation from $\mathcal{D}^u$.
Because these non-smooth samples fail to acquire enough label information, the task model has a vague understanding of the local distribution of these samples.
In this sense, selecting non-smooth samples for further annotation are more valuable, which can help the task model smooth unlabeled samples better and output more stable predictions.

To select these non-smooth samples, the virtual inconsistency ranker estimates the inconsistency of the task model’s prediction on unlabeled samples under augmentation strategies of two different granularity.
Specifically, for each unlabeled sample $\textbf{x}_i^u$ in the same unlabeled pool as mentioned in section ~\ref{sec:slp} (this section, we use a different symbol $\textbf{x}_i^u$ to distinguish between the selection stage and the SSL training stage):
(1) we first obtain its coarse-grained augmentation set $\bar{\textbf{X}}^u_i=\{\bar{\textbf{x}}^u_{i,1}, \dots, \bar{\textbf{x}}^u_{i,K}\}$ by aforementioned augmentations, and feed $\textbf{x}_i^u$ and $\bar{\textbf{X}}^u_i$ into the task model to get their predictions $\bar{y}_i^u$ and $\bar{\textbf{Y}}^u_i =\{\bar{y}_{i,1}^u, \dots,  \bar{y}_{i,K}^u\}$.
(2) Then we feed $\bar{\textbf{X}}^u_i$ and $\bar{\textbf{Y}}^u_i$ simultaneously into the ranker to get adversarial perturbation $\textbf{R}_i^{adv}= \{\textbf{r}_{i,1}^{adv}, \dots, \textbf{r}_{i,K}^{adv}\}$ for each augmented sample $\bar{\textbf{x}}_{i,k}^u \in \bar{\textbf{X}}^u_i$.
(3) After that, we feed each fine-grained augmented sample $\hat{\textbf{x}}_{i,k}^u= \bar{\textbf{x}}^u_{i,k}+\textbf{r}_{i,k}^{adv}$ into the task model again to get their fine-grained augmented predictions $\hat{\textbf{Y}}^u_i =\{\hat{y}_{i,1}^u, \dots,  \hat{y}_{i,K}^u\}$. Besides, we conduct the fine-grained augmentations on texts’ embedding vectors extracted from the middle layer of the task model, as texts’ input space is discrete. 
The process of generating adversarial perturbation is formulated as:
\begin{equation}
\textbf{r}^{adv} = \mathop{\arg \max}_{\Delta \textbf{r},||\Delta \textbf{r}||\leq \epsilon} \mbox{KL}(p(\bar{y}^u\arrowvert \bar{\textbf{x}}^{u},\theta), p(\hat{y}^u\arrowvert \bar{\textbf{x}}^{u} + \Delta \textbf{r},\theta))\label{r1}. 
\end{equation}
where $p(y\arrowvert \textbf{x},\theta)$ represents the posterior probability of the task model. Under perturbation of the same norm $\epsilon$, there is a higher probability for adversarial samples of unlabeled samples with unstable predictions to change their original label and obtain predictions of other classes. Therefore, the ranker computes the KL-divergence of the posterior probability of samples and their adversarial samples to measure the inconsistency of unlabeled samples.

Since the computation of $\textbf{r}^{adv}$ is intractable for many neural networks,~\cite{miyato2018virtual} proposed to approximate $\textbf{r}^{adv}$ using the second-order Taylor approximation and solved the $\textbf{r}^{adv}$ via the power iteration method. Specifically, we can approximate $\textbf{r}^{adv}$ by applying the following update:

\begin{equation}
    \textbf{r}^{adv} \leftarrow \epsilon \overline{\nabla_{\Delta \textbf{r}}
    \mbox{KL}(p(\bar{y}^u|\bar{\textbf{x}}^u), p(\hat{y}^u|\bar{\textbf{x}}^u + \Delta \textbf{r}))}\label{r2}.
\end{equation}
where $\Delta \textbf{r}$ is a randomly sampled unit vector, $p(\bar{y}^u|\bar{\textbf{x}}^u)$ is the label for coarse-grained augmented samples, $p(\hat{y}^u|\bar{\textbf{x}}^u + \Delta \textbf{r})$ is the fine-grained augmented prediction, and the sign $\overline{\textbf{v}}$ means  the unit vector of $\textbf{v}$. The computation of $\nabla_{\Delta \textbf{r}}\mbox{KL}$ can be performed with one set of backpropagation for neural networks. 

Once the adversarial perturbation $\textbf{r}^{adv}$ is solved, we can estimate the total inconsistency ~\cite{gao2020consistency} of unlabeled samples under coarse-grained and fine-grained augmentations. Coarse-grained inconsistency can be formulated as:

\begin{equation}
\begin{aligned}
    {In}_{coa}(\textbf{x}^u_i)=\sum_{c=1}^{C}Var & \lbrack p(\bar{y}_{i}^u=c\arrowvert \textbf{x}_i^u,\theta), p(\bar{y}_{i,1}^u=c\arrowvert \textbf{x}_{i,1}^u,\theta), \\ 
    &\dots, p(\bar{y}_{i,K}^u=c\arrowvert \textbf{x}_{i,K}^u,\theta)\rbrack.
\end{aligned}
\label{adv}
\end{equation}
where $C$ is the number of classes, and $Var$ means the variance. Furthermore, fine-grained inconsistency can be formulated as:

\begin{equation}
    {In}_{fin}(\textbf{x}^u_i)=\sum_{k=1}^{K}KL(\bar{y}^u_{i,k},\hat{y}^u_{i,k}). 
\label{adv}
\end{equation}
where $KL$ means the KL-divergence.
To combine the above two selection criteria together, we normalize and transform them into percentiles.
Specially, we denote $\Phi{\varphi}(\textbf{x}^u_i,\mathcal{D}^u)$ as the percentile of the criteria $\varphi$ of a given sample among the dataset $\mathcal{D}^u$.
For instance, $\Phi_{\varphi}(\textbf{x}^u_i,\mathcal{D}^u)=75\%$ indicates that $75\%$ of the samples in $\mathcal{D}^u$ are smaller than $\textbf{x}^u_i$ under the criteria $\varphi$.
We can calculate the total inconsistency of each unlabeled sample as:

\begin{equation}
    {In}(\textbf{x}^u_i)=\gamma\cdot\Phi_{{In}_{coa}}(\textbf{x}^u_i,\mathcal{D}^u)+(1-\gamma)\cdot\Phi_{{In}_{fin}}(\textbf{x}^u_i,\mathcal{D}^u).
\label{adv}
\end{equation}
where $\gamma$ is the weight coefficient to balance two criteria.
Finally, we select top-$\mathcal{M}$ samples with the largest inconsistency ${In}(\textbf{x}^u_i)$ from unlabeled samples as the initial annotation candidates.

\subsection{Density-aware Uncertainty Re-ranker}\label{sec:bl}

Based on the goal described in section~\ref{sec:vad}, the virtual inconsistency ranker roughly selects annotation candidates with large inconsistency, which can be regarded as an initial recall set of final potential samples.
However, there is still a gap between the goal of the virtual inconsistency ranker (\textit{i.e.} select samples with large inconsistency) and our final AL goal (\textit{i.e.} select samples with high uncertainty).
Here, we propose a density-aware uncertainty re-ranker to rank the annotation candidates further, guaranteeing that the selected samples are with high uncertainty and large inconsistency.
In this way, the top samples in the candidates set can bring great model uncertainty reduction.

In practice, we use the entropy of predicted labels from the task model to estimate the uncertainty of samples and select top-$\mathcal{K}$ samples from former top-$\mathcal{M}$ annotation candidates.
The entropy of $i$-th unlabeled sample $\textbf{x}_i^m$ from the set $\mathcal{M}$ can be calculated with the following formulation:
\begin{equation} 
    {En}^{'}(\textbf{x}^m_i) = -\sum_{c}^{C} P(y_c|\textbf{x}^m_i) \, \log P(y_c|\textbf{x}^m_i).
\end{equation}
where $P(y_c|\textbf{x}^m_i)$ is the probability of the $i$-th unlabeled sample in the set $\mathcal{M}$ belonging to the $c$-th class.

However, the above entropy-based formula only estimates the uncertainty information of each sample and fails to take the distribution relations among samples into account.
As a result, the metric may run the risk of selecting some outliers or unrepresentative samples in the distribution space.
To alleviate this issue, we re-weight the uncertainty metric with a representativeness factor and explicitly consider the distribution structure of samples. We denote this density-aware uncertainty as:
\begin{equation} 
\begin{aligned}
    {En}(\textbf{x}^m_i) = {En}^{'}(\textbf{x}^m_i)\times{\left(\frac{1}{M}\sum_{\textbf{x}^{'}\in\mathcal{M}}sim(\textbf{x}^m_i,\textbf{x}^{'})\right)}.
\end{aligned}
\end{equation}
where $M$ is the size of the candidates set $\mathcal{M}$, and the additional second term in the formula is the similarity of each sample $\textbf{x}_i^m$ relative to other samples in the distribution space.
We can obtain the similarity term through cosine similarity, Euclidean distance, Spearman’s rank correlation, or any other metrics. Considering that cosine similarity is common to compute the similarity of texts’ embedding and also effective for images’ similarity computation, we use cosine similarity in our paper:

\begin{equation} 
    sim(\textbf{x}^m_i,\textbf{x}^{'}) = \frac{{\textbf{x}^m_i}^T \cdot {\textbf{x}^{'}}}{||\textbf{x}^m_i|| \times ||\textbf{x}^{'}||}.
\end{equation}
and this density-aware indicator can help us select samples with high uncertainty as well as spatial representativeness.

Accordingly, SSL and AL can work collaboratively for reciprocal enhancement in this framework. On the one hand, the SSL label propagator can propagate label information of labeled samples and smooth the local distribution around unlabeled samples through a mixture of labeled and unlabeled samples. Then SSL can provide a consistent signal to help AL decide the annotation candidates.
On the other hand, the virtual inconsistency ranker and the density-aware uncertainty re-ranker enable human to annotate the samples with the largest inconsistency plus the highest uncertainty weighted by representativeness. AL assists the SSL label propagator exclude unstable and unsmoothed augmentation samples for better label propagation. We detail IDEAL in Algorithm~\ref{alg} in Appendix~\ref{alg_sec}.

\section{Experiments}
To validate the proposed method, we conduct a comprehensive evaluation in both text and image domains, with four benchmark datasets and two new real-world datasets. We initiate each task with a small (randomly sampled) labeled training set, and the remaining unlabeled data are used as candidates for IDEAL/baseline's AL/SSL components. Once labeled by human, samples will be added to the training set. For each cycle, we train the task model with the newly updated training set. For a fair comparison, all baselines share the same initial training set and model parameters. All the figures in this study depict the results over five experiment trials.\footnote{To facilitate other scholars reproduce the experiment outcomes, we will make all the data code available via https://github.com/AmazingJeff/IDEAL.} 

\subsection{Datasets}
We choose two text classification benchmarks: AG News~\cite{zhang2015character} and IMDB~\cite{maas2011learning}, and two image classification benchmarks: CIFAR-10~\cite{krizhevsky2009learning} and CIFAR-100~\cite{krizhevsky2009learning}. Furthermore, we collect two industrial text classification datasets. The Legal Text dataset is collected from Chinese court trial cases, used to classify and retrieve similar cases. We define 12 fact labels (elements of judgment basis) for each case. The Bidding dataset is used to facilitate the users to filter the procurement methods (different companies prefer different procurement methods) and help them locate the most suitable bidding opportunities. We define 22 labels of purchase and sale for each announcement document. Unlike benchmarks, Legal Text and Bidding datasets come from industrial applications, and their annotations need experts with domain knowledge. The average annotation price of each sample for these two datasets is $\$$0.91 and $\$$0.92 respectively. 

%
%

\subsection{Baselines}
For text classification tasks, we employ four baselines for comparison:
(1) EGL~\cite{zhang2017active}; (2) Core-set~\cite{sener2017active}; (3) MC-dropout~\cite{siddhant2018deep};
(4) Random sampling~\cite{figueroa2012active}.
For supervised image classification tasks, we adopt seven baselines including: (1) Core-set~\cite{sener2017active}; (2) Random sampling~\cite{figueroa2012active}; (3) ICAL~\cite{gao2020consistency}; (4) SRAAL~\cite{zhang2020state}; (5) VAAL~\cite{sinha2019variational}; (6) LLAL~\cite{yoo2019learning}; (7) REVIVAL~\cite{guo2021semi}. Since some baselines under supervised learning can not be applied or transferred to SSL scenarios directly, we use different baselines
in the SSL scenario. The baseline details can be found in Appendix~\ref{baseline}. 


\subsection{Evaluation Settings}
We adopt Wide ResNet-28~\cite{oliver2018realistic} as the backbone of task models for image classification, while we use a BERT model along with a two-layer MLP architecture as text tasks' backbone. The initial training set is uniformly distributed over classes, and we set up the initial set by randomly sampling. Under the supervised learning scenario,
we only use the labeled data to train the task model. Thus, we leave out the operations of the SSL propagator. Under the SSL scenario, SSL hyper-parameters have been well-explored in MixText~\cite{chen2020mixtext} and Mixmatch~\cite{berthelot2019mixmatch}. They found in practice that hyper-parameters can be fixed and do not need to be tuned on a 
per-experiment or per-dataset basis. Thus, we follow these empirical settings. For text, we set K to 2 (back translation from German and Russian). Based on translation qualities, the corresponding weight parameters ($w_u$, $w_{Ger}$, $w_{Ru}$) for original samples and their augmented samples are (1, 1, 0.5). Over real-world datasets, we set back translation (English and German) weight parameters ($w_u$, $w_{En}$, $w_{Ger}$) to (1, 0.8, 0.2). For images, we set K to 5 (5 augmentations obtained by horizontally flipping and random cropping) and all weight parameters $w$ to 1. We set $\alpha$ to 16 and 0.75 for text and image datasets respectively. Further, we refer to the work~\cite{miyato2018virtual} to fine-tune the hyper-parameter $\epsilon$, and we obtain $\epsilon$ as 1e-2 and 10 for text and image data respectively.

\subsection{Performance Analysis}
\begin{figure*}[htbp]
\centering
\subfigure[CIFAR-10.]{

\begin{minipage}[t]{0.24\linewidth}
\centering
\label{cifar10}
\includegraphics[width=1.68in]{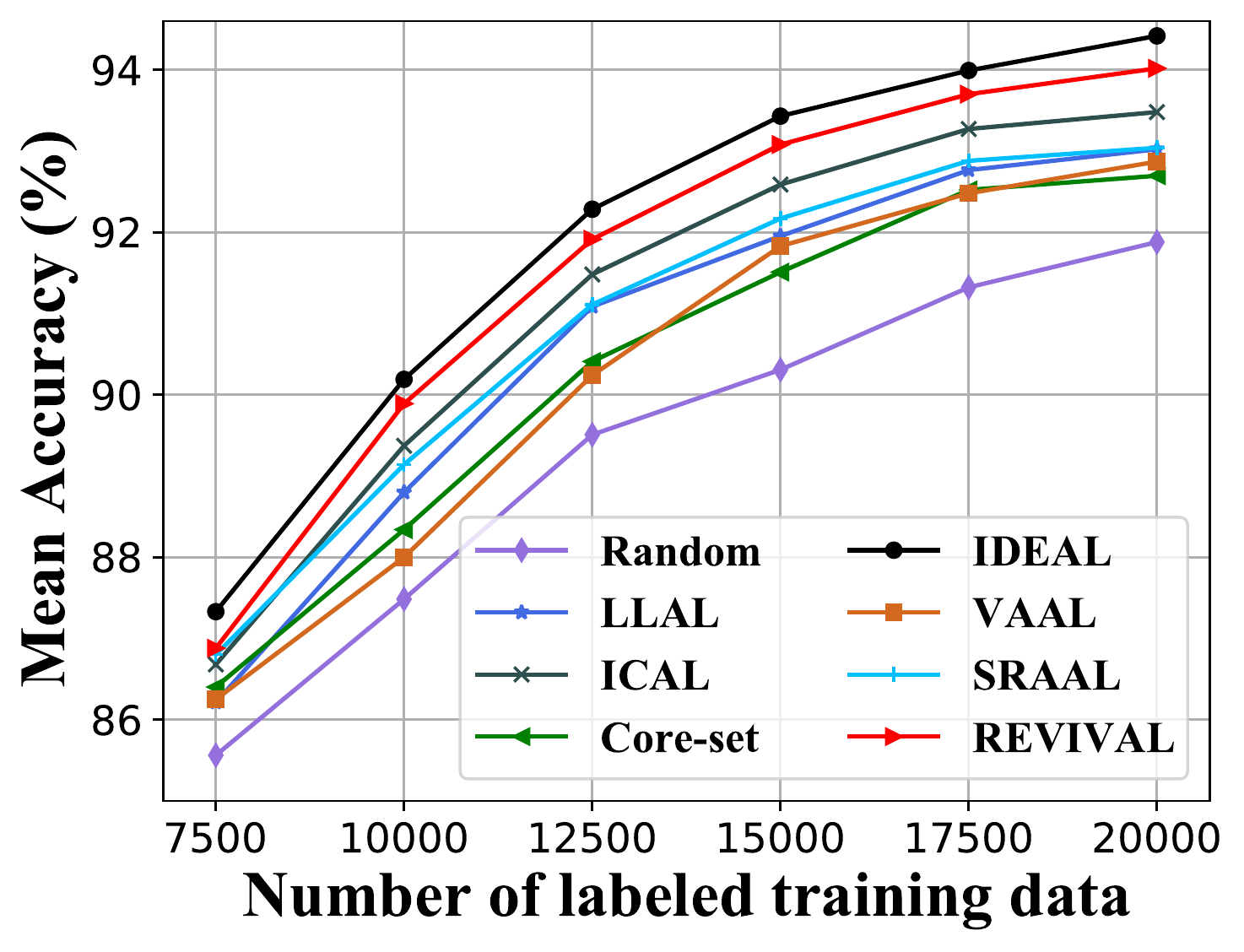}

\end{minipage}%
}%
\subfigure[CIFAR-100.]{
\begin{minipage}[t]{0.24\linewidth}
\centering
\label{cifar100}
\includegraphics[width=1.68in]{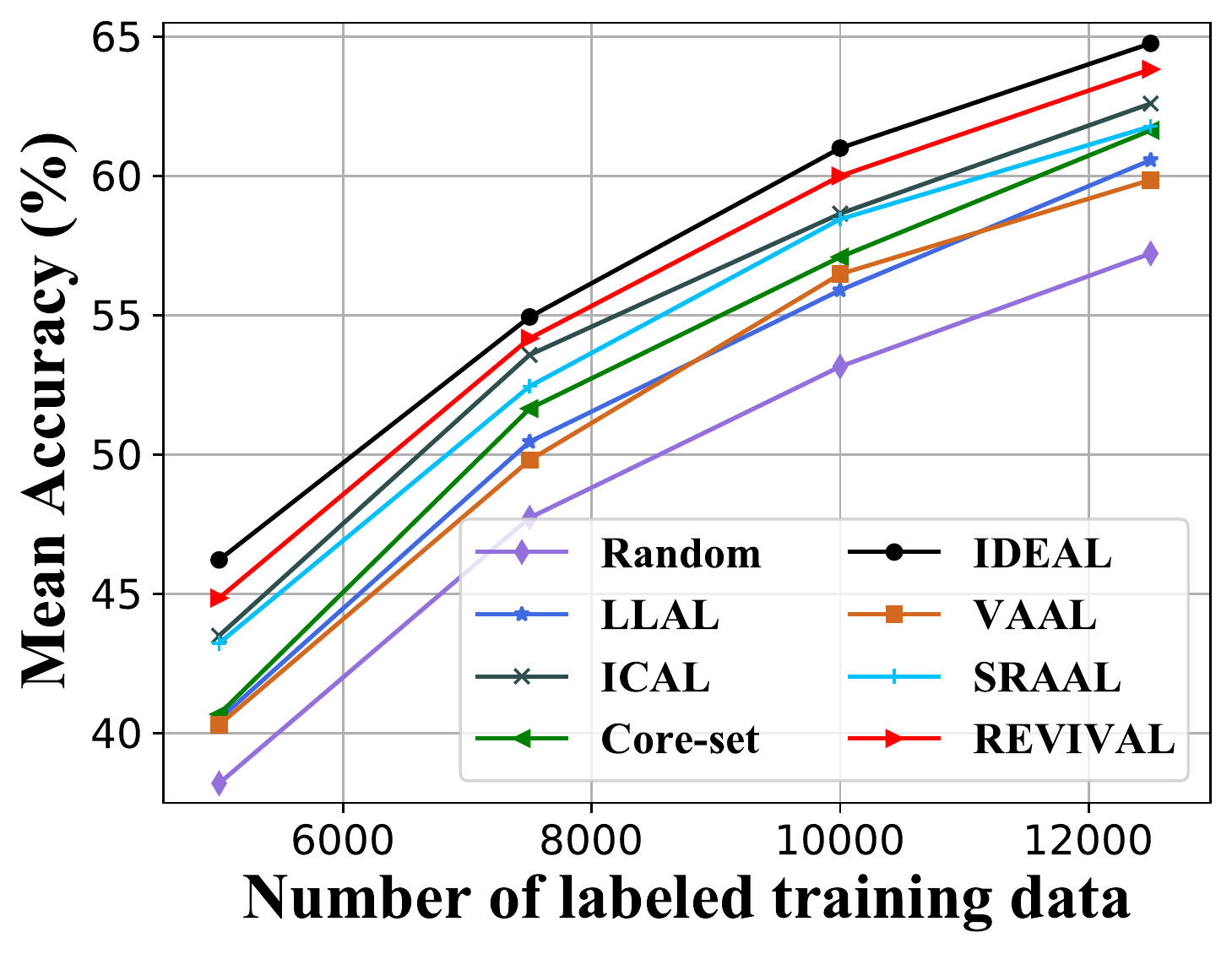}

\end{minipage}%
}%
\subfigure[CIFAR-10.]{
\begin{minipage}[t]{0.24\linewidth}
\centering
\label{mix_cifar10}
\includegraphics[width=1.68in]{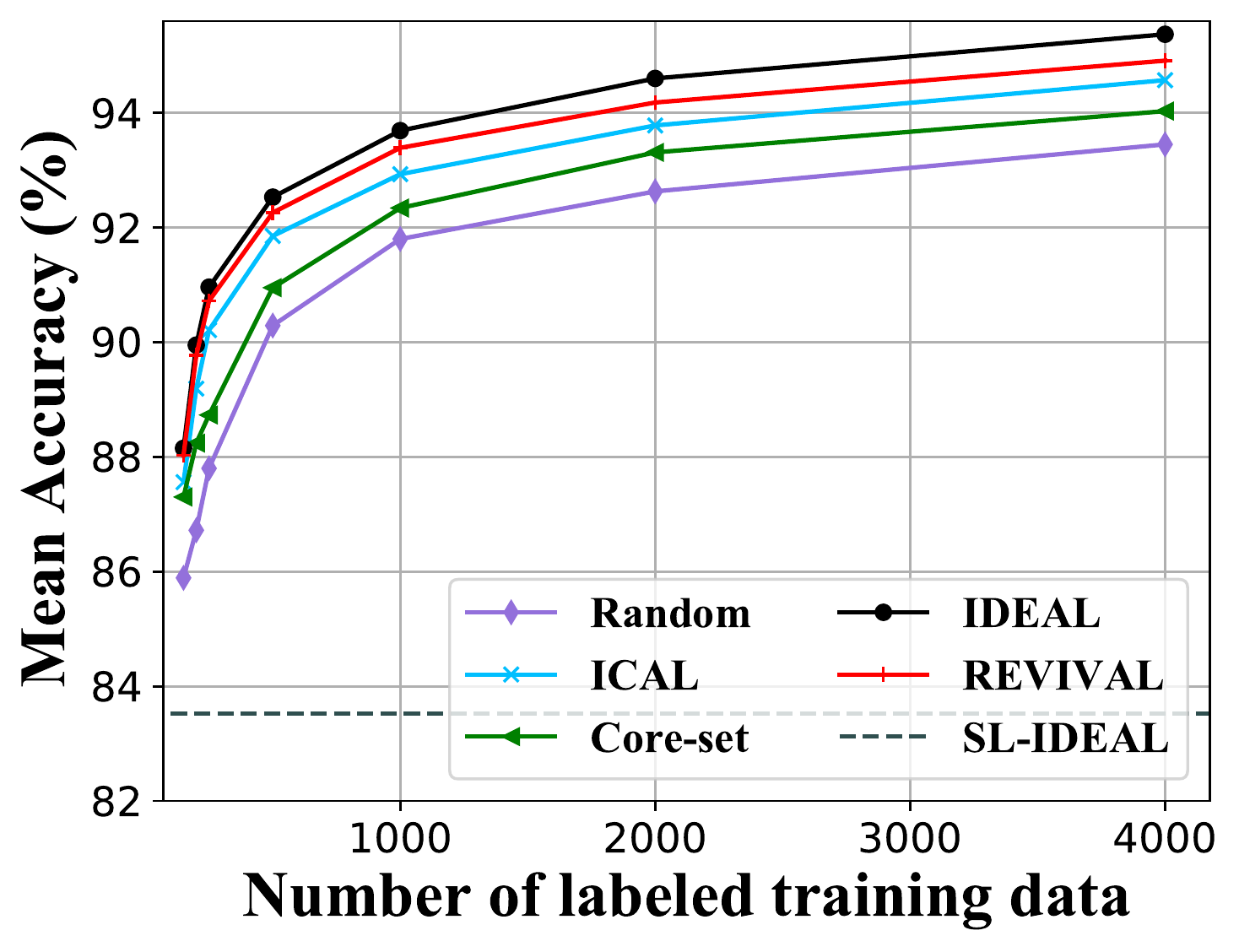}

\end{minipage}
}%
\subfigure[CIFAR-100.]{
\begin{minipage}[t]{0.24\linewidth}
\centering
\label{mix_cifar100}
\includegraphics[width=1.68in]{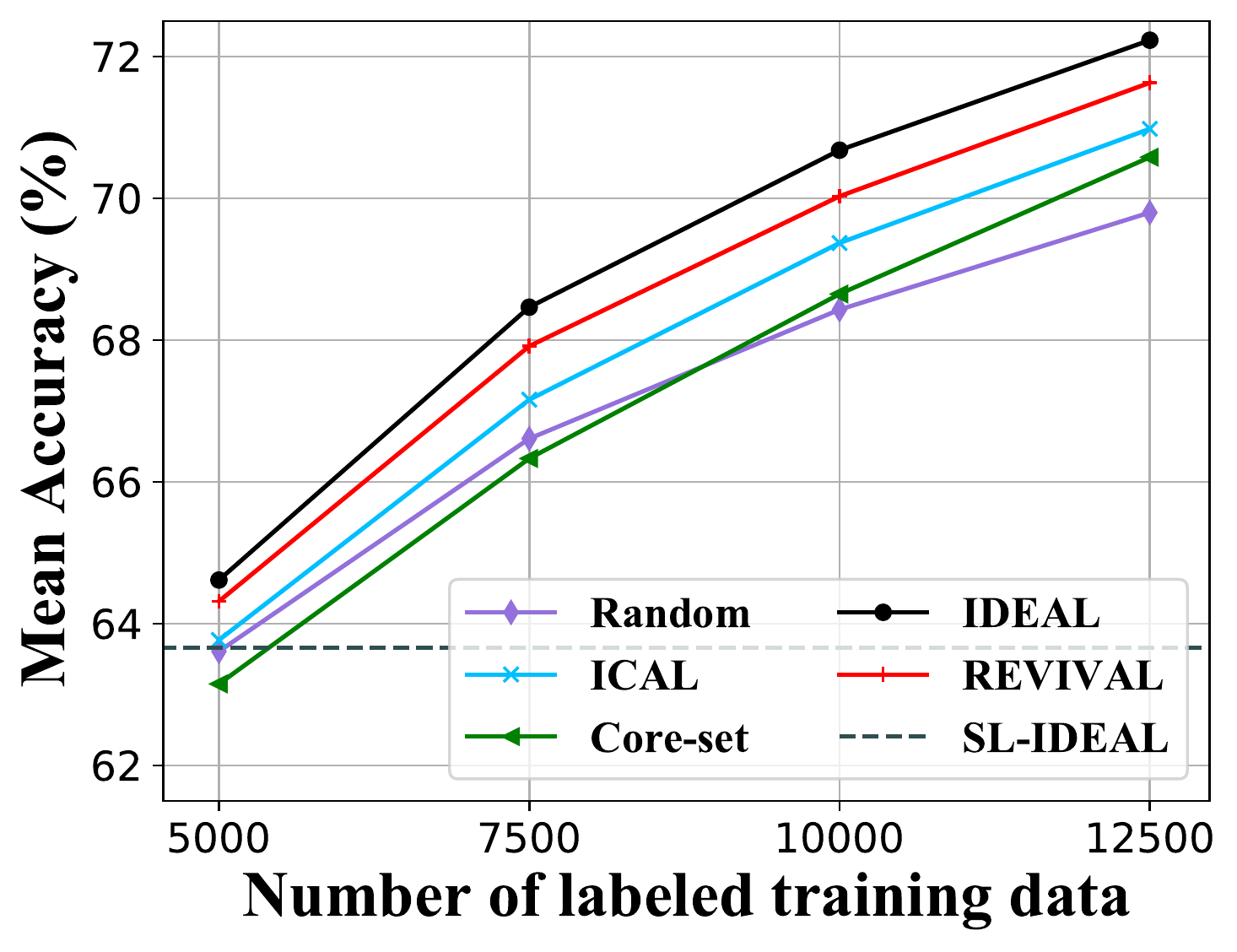}

\end{minipage}
}%
\centering
\vspace{-0.5cm}
\caption{Performance over image benchmarks. Results in (a) and (b) are obtained under supervised learning. Results in (c) and (d) are obtained under semi-supervised learning. The dotted lines (SL-IDEAL) at the bottom in (c) and (d) represent the performance of IDEAL under supervised learning with 4000 and 12500 labeled data respectively.}
\vspace{-0.3cm}
\end{figure*}

\noindent $\bullet$ \textbf{Performance for image classification.}\quad 
Figures~\ref{cifar10} and~\ref{cifar100} depict the supervised learning performance of IDEAL on image benchmarks. For CIFAR-10~(Figure \ref{cifar10}), we can observe that IDEAL outperforms evidently against state-of-the-art methods in all selection cycles. In Figure~\ref{cifar100}, IDEAL beats all baselines with \textbf{a margin up to} $\textbf{1.37\%}$. 
Figures ~\ref{mix_cifar10} and ~\ref{mix_cifar100} show that, with the SSL setting, IDEAL is superior to baselines throughout the entire sample selection process. Compared to AL, SSL brings essential improvements early in the selection process, so it is necessary to have different selection batch sizes for better visualization of the performance increase under the SSL scenario. 
In Figure~\ref{mix_cifar10}, the accuracy of methods under the SSL scenario (with 150 labeled samples) is higher than that under the supervised learning scenario (with 4000 labeled samples). IDEAL reaches $94.6\%$ accuracy with 2000 labeled samples, while previous SOTA requires more than 3500 (IDEAL significantly \textbf{reduces the annotation cost by at least} $\textbf{42.8\%}$). Compared to mere SSL (random selection), IDEAL brings a huge reduction in annotation cost of $75\%$ (4000→1000) to reach $93.69\%$ accuracy.
In Figure~\ref{mix_cifar100}, when using 12500 labeled samples, semi-supervised IDEAL achieves $8.57\%$ more accuracy than supervised IDEAL. 

\begin{figure*}[htbp]
\centering
\subfigure[AG News.]{
\begin{minipage}[t]{0.24\linewidth}
\centering
\label{ag}
\includegraphics[width=1.66in]{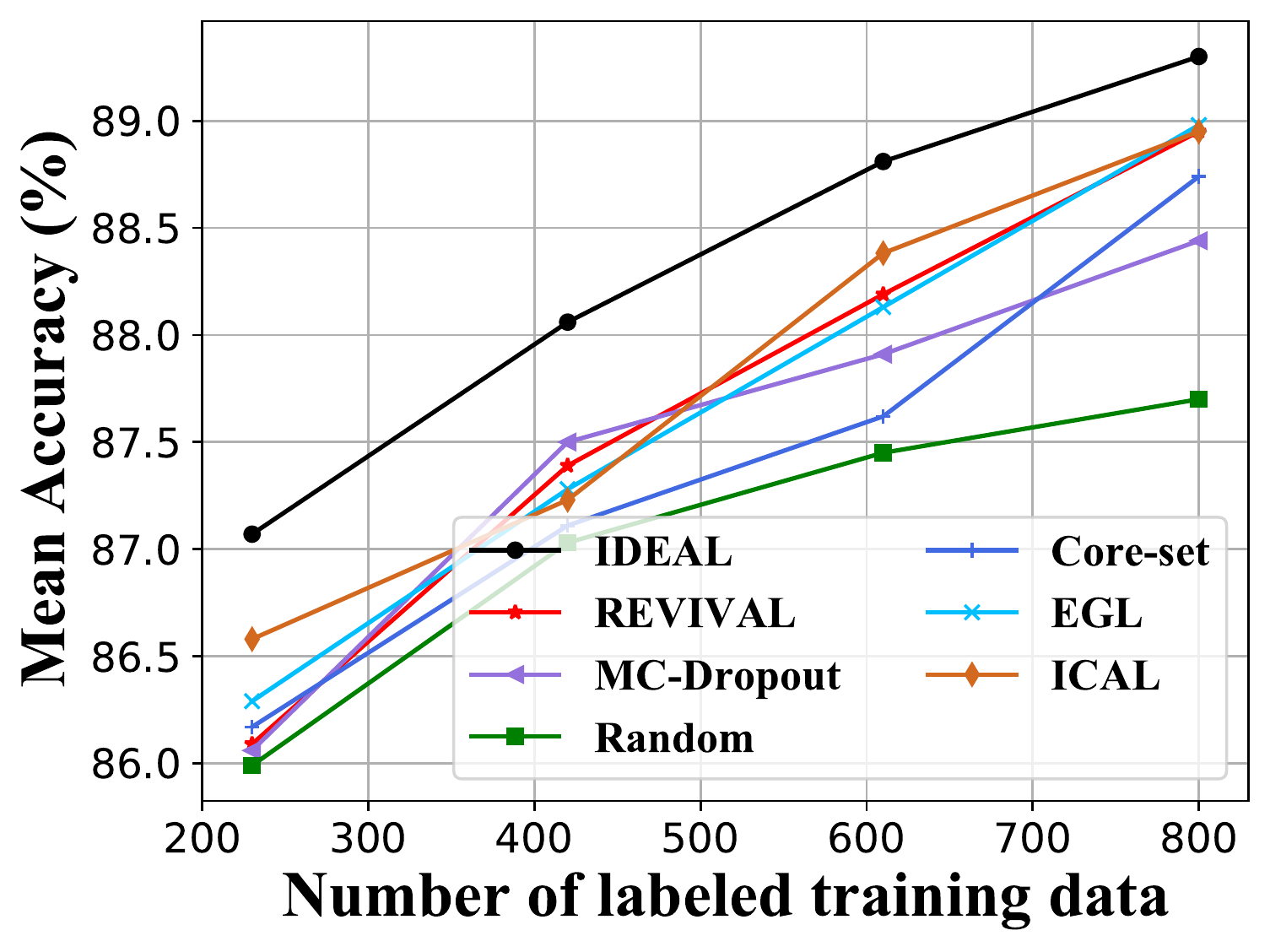}

\end{minipage}%
}%
\subfigure[IMDB.]{
\begin{minipage}[t]{0.24\linewidth}
\centering
\label{imdb}
\includegraphics[width=1.66in]{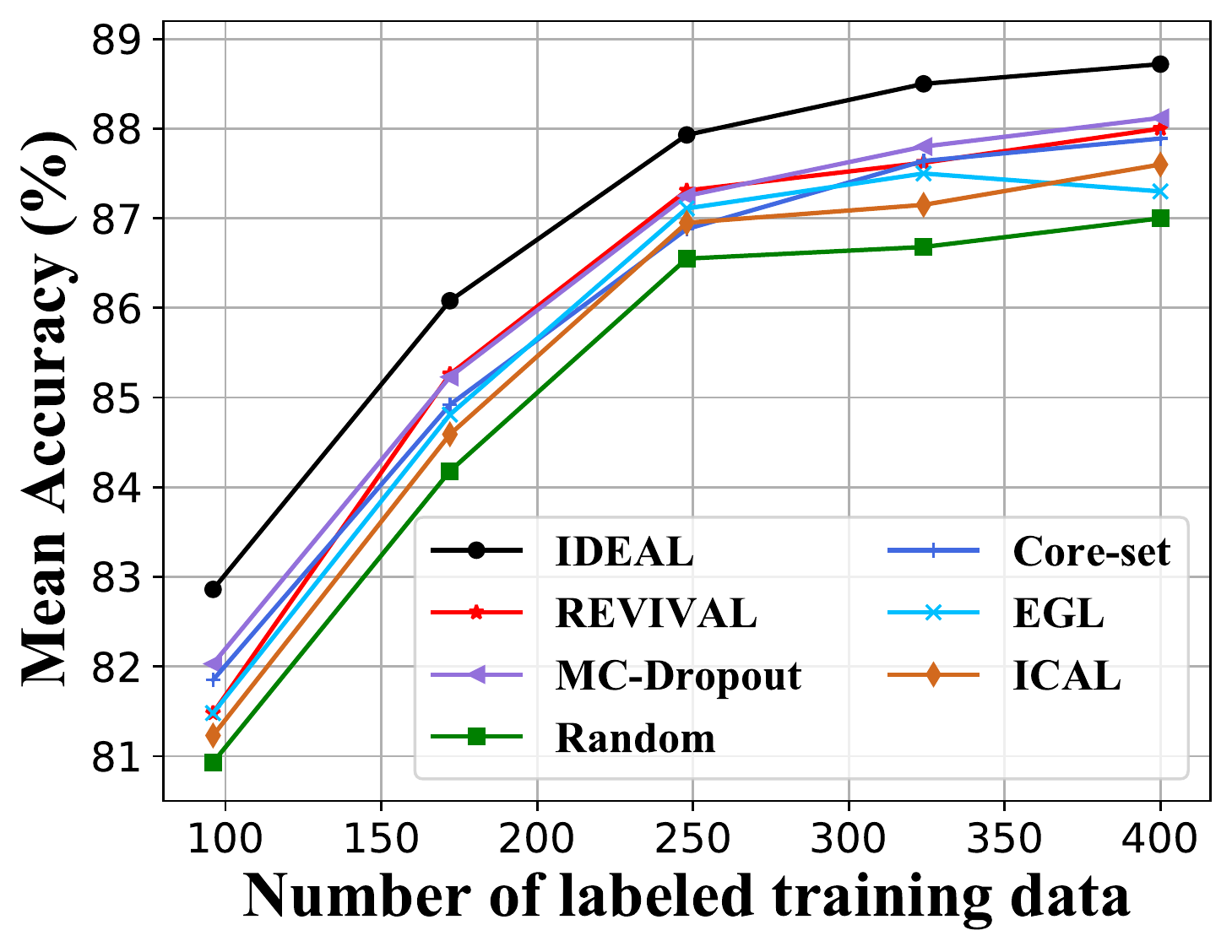}

\end{minipage}%
}%
\subfigure[AG News.]{
\begin{minipage}[t]{0.24\linewidth}
\centering
\label{semi-ag}
\includegraphics[width=1.66in]{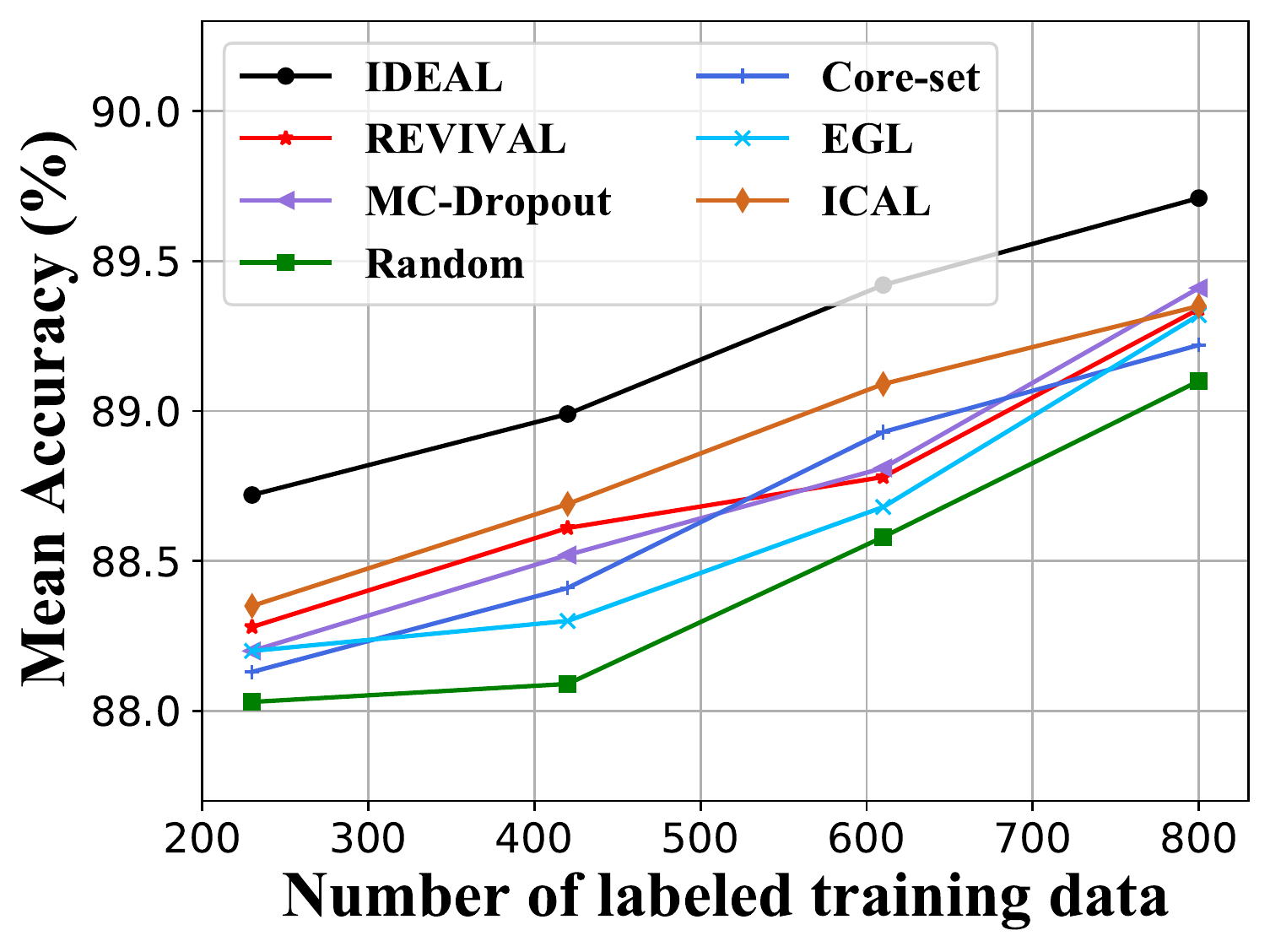}

\end{minipage}
}%
\subfigure[IMDB.]{
\begin{minipage}[t]{0.24\linewidth}
\centering
\label{semi-imdb}
\includegraphics[width=1.66in]{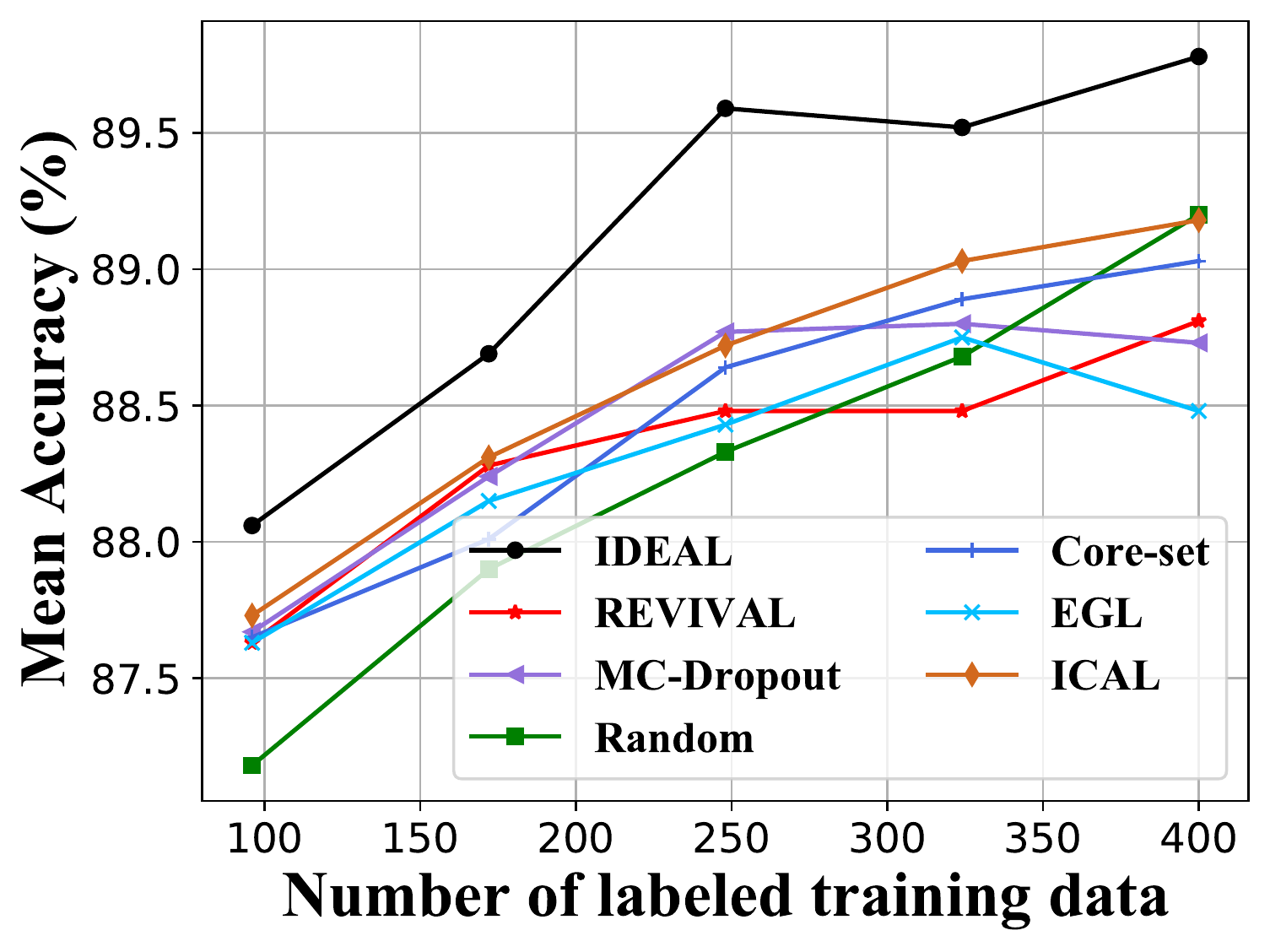}

\end{minipage}
}%
\centering
\vspace{-0.5cm}
\caption{Performance over text benchmarks. Results in (a) and (b) are obtained under supervised learning. Results in (c) and (d) are obtained under semi-supervised learning.}
\vspace{-0.3cm}
\end{figure*}

\noindent $\bullet$ \textbf{Performance for text classification.}\quad 
Figures~\ref{ag} and~\ref{imdb} show the results of supervised IDEAL for text classification.
One can witness that IDEAL \textbf{outperforms all baselines} throughout the whole sample selection stage. On the `BERT scale', IDEAL leads to decent performance improvement, which validates the effectiveness of the proposed method. Figures ~\ref{semi-ag} and ~\ref{semi-imdb} show the results of semi-supervised IDEAL for the text classification task, and the Random represents the text SSL method MixText (w/o AL). From the figures, we can see that IDEAL \textbf{consistently demonstrates the best performances} across different labeled data numbers in both datasets. The graph propagator limits REVIVAL's performance by failing to capture complex semantic relations among text samples and build a high-quality relation graph. 
In contrast, IDEAL leverages the Mixup technique to propagate label information for subsequent inconsistency estimation. Further, IDEAL can not only measure the prediction inconsistency of augmented samples but estimates the robustness of samples' embedding to adversarial perturbation as well. Thus, IDEAL significantly improves the naive SSL method (the Random) by excluding unsmoothed and unstable samples, compared to ICAL and REVIVAL.  

\vspace{-0.6cm}
\begin{table}[h]
\small
\caption{ \textbf{Annotation cost comparison  with similar accuracy on Legal and Bidding datasets. Superscript $^{\dagger}$  indicates the SOTA SSL-AL model.  $\downarrow$ suggest that the small value is better. }}
\vspace{-0.2cm}
\centering\setlength{\tabcolsep}{1.2mm}{
\begin{tabular}[width=1\textwidth]{l|cccc|c|c}
\toprule[1.5pt]
\multicolumn{1}{c|}{{\textbf{Methods}}} & \multicolumn{1}{c|}{\textbf{Dataset}}&  \multicolumn{1}{|c}{\textbf{Setting}} & \multicolumn{1}{c}{\textbf{N}} & \multicolumn{1}{c|}{\textbf{$\textbf{\$}$/$\textbf{L}$}}  & \multicolumn{1}{c|}{ \textbf{Cost $\textbf{\$}$ }$\downarrow$} & \multicolumn{1}{c}{\textbf{Accuracy \%}} 
\\\toprule[1pt]
Random~\cite{figueroa2012active}&\multicolumn{1}{c|}{\multirow{3}{*}{Legal}} &SSL	&9,000 &0.91 &8,190 &87.71		\\ \cmidrule(l){1-1} \cmidrule(l){3-7}
ICAL~\cite{gao2020consistency}$^{\dagger}$  &\multicolumn{1}{c|}{} &SSL-AL	&7,000 &0.91  &6,370&87.77		\\\cmidrule(l){1-1} \cmidrule(l){3-7}
\textbf{IDEAL} &\multicolumn{1}{c|}{}&\textbf{SSL-AL}	&\textbf{5,000}&\textbf{0.91}  &\textbf{4,550}&\textbf{88.36}	\\

\midrule[1pt]
Random~\cite{figueroa2012active} &\multicolumn{1}{c|}{\multirow{2}{*}{Bidding}} &SSL	&2,500 &0.92 &2,300	&{85.35}	\\ \cmidrule(l){1-1} \cmidrule(l){3-7}
\textbf{IDEAL} &\multicolumn{1}{c|}{}&\textbf{SSL-AL}	&\textbf{1,000} &\textbf{0.92}  &\textbf{920}&\textbf{86.25}\\\midrule[1pt]
ICAL~\cite{gao2020consistency}$^{\dagger}$  &\multicolumn{1}{c|}{\multirow{2}{*}{Bidding}}&SSL-AL	&2,500 &0.92  &2,300	&{87.60}	\\\cmidrule(l){1-1} \cmidrule(l){3-7}
\textbf{IDEAL} &\multicolumn{1}{c|}{}&\textbf{SSL-AL}	&\textbf{2,000} &\textbf{0.92}  &\textbf{1,840}&\textbf{87.70}	

\\ \bottomrule[1.5pt] 
\end{tabular}}
\label{tab:selected_data}
\vspace{-0.6cm}
\end{table}

\begin{figure*}[htbp]
\centering
\subfigure[Legal Text.]{
\begin{minipage}[t]{0.24\linewidth}
\centering
\label{law}
\includegraphics[width=1.68in]{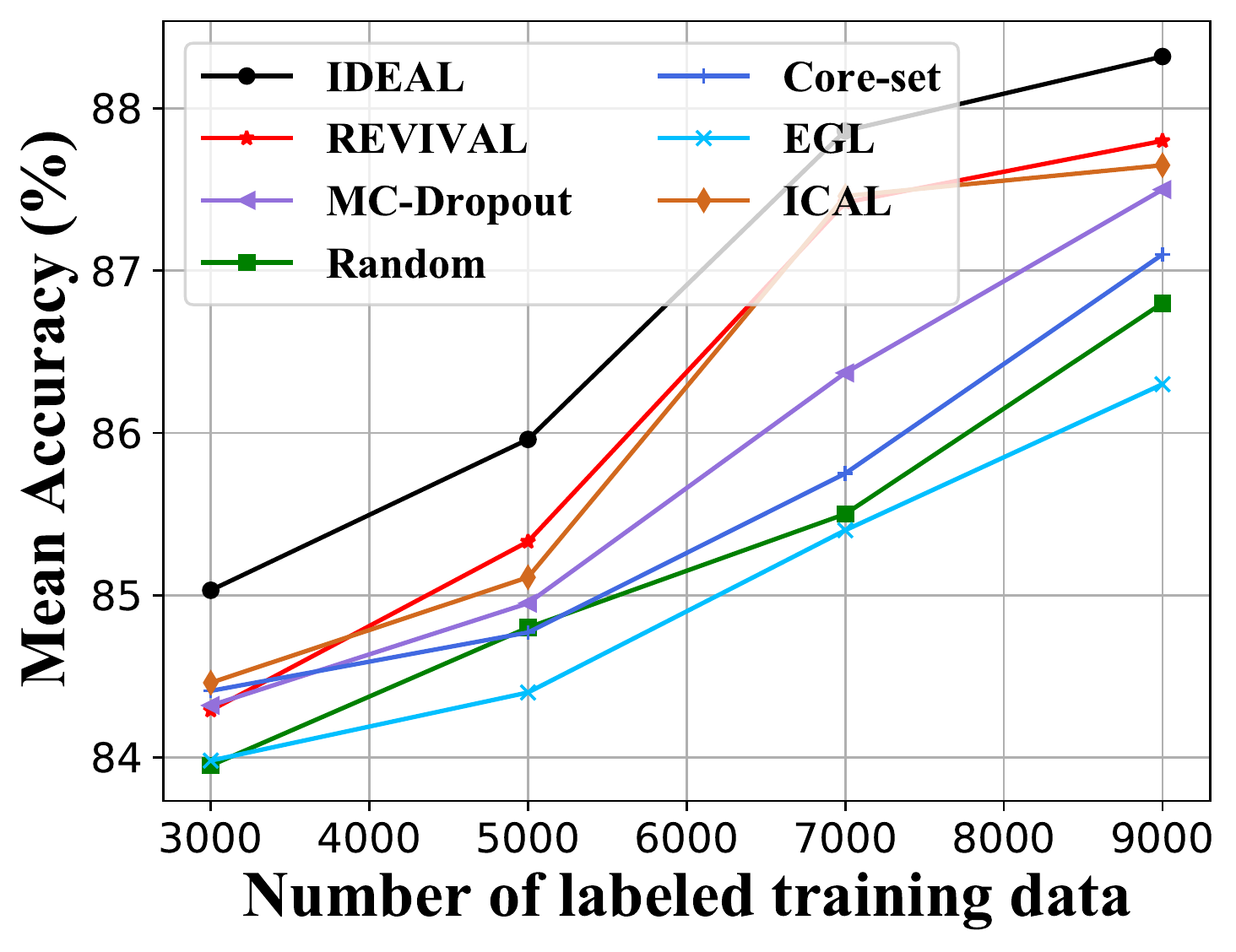}

\end{minipage}%
}%
\subfigure[Bidding.]{
\begin{minipage}[t]{0.24\linewidth}
\centering
\label{bid}
\includegraphics[width=1.68in]{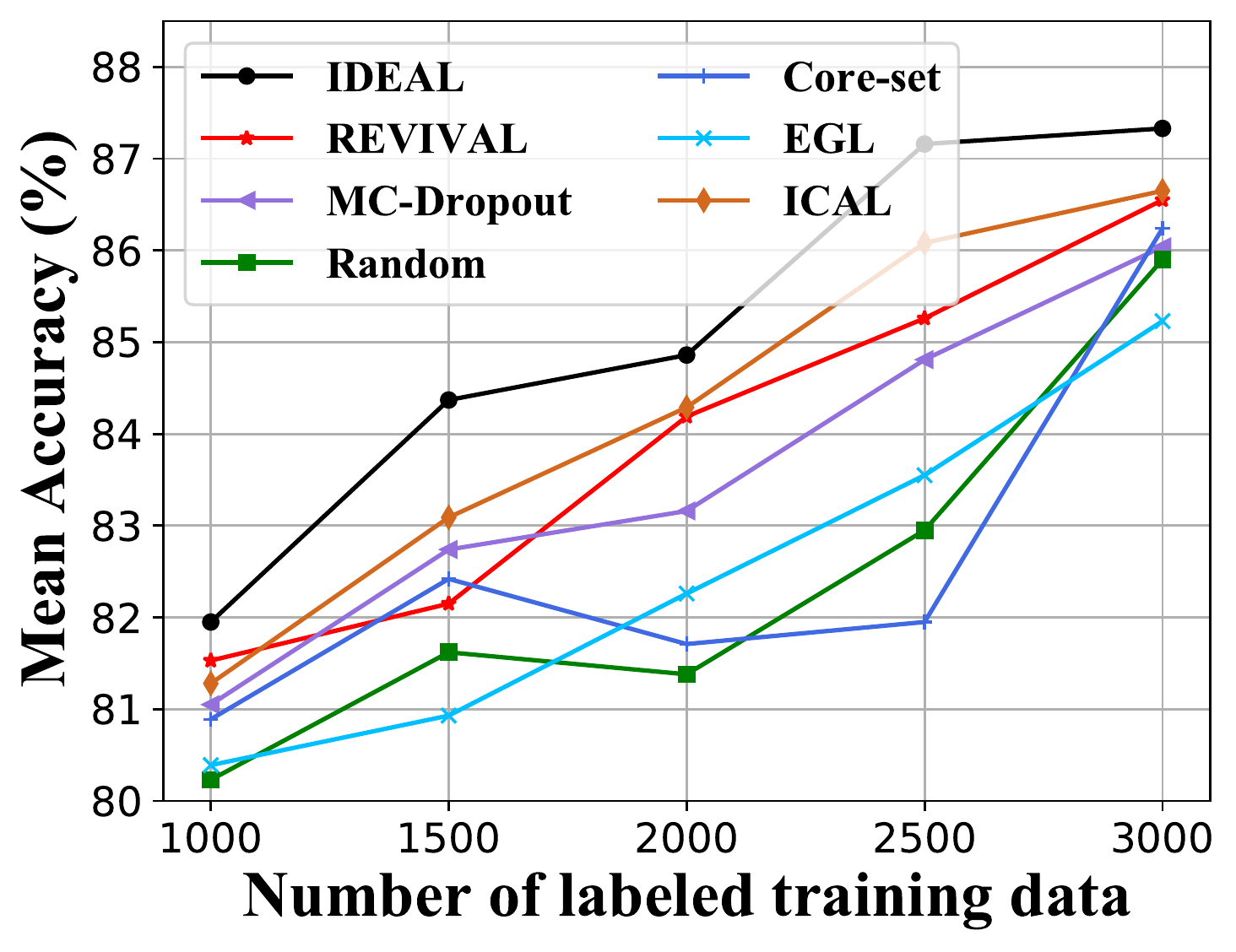}

\end{minipage}%
}%
\subfigure[Legal Text.]{
\begin{minipage}[t]{0.24\linewidth}
\centering
\label{semi-law}
\includegraphics[width=1.68in]{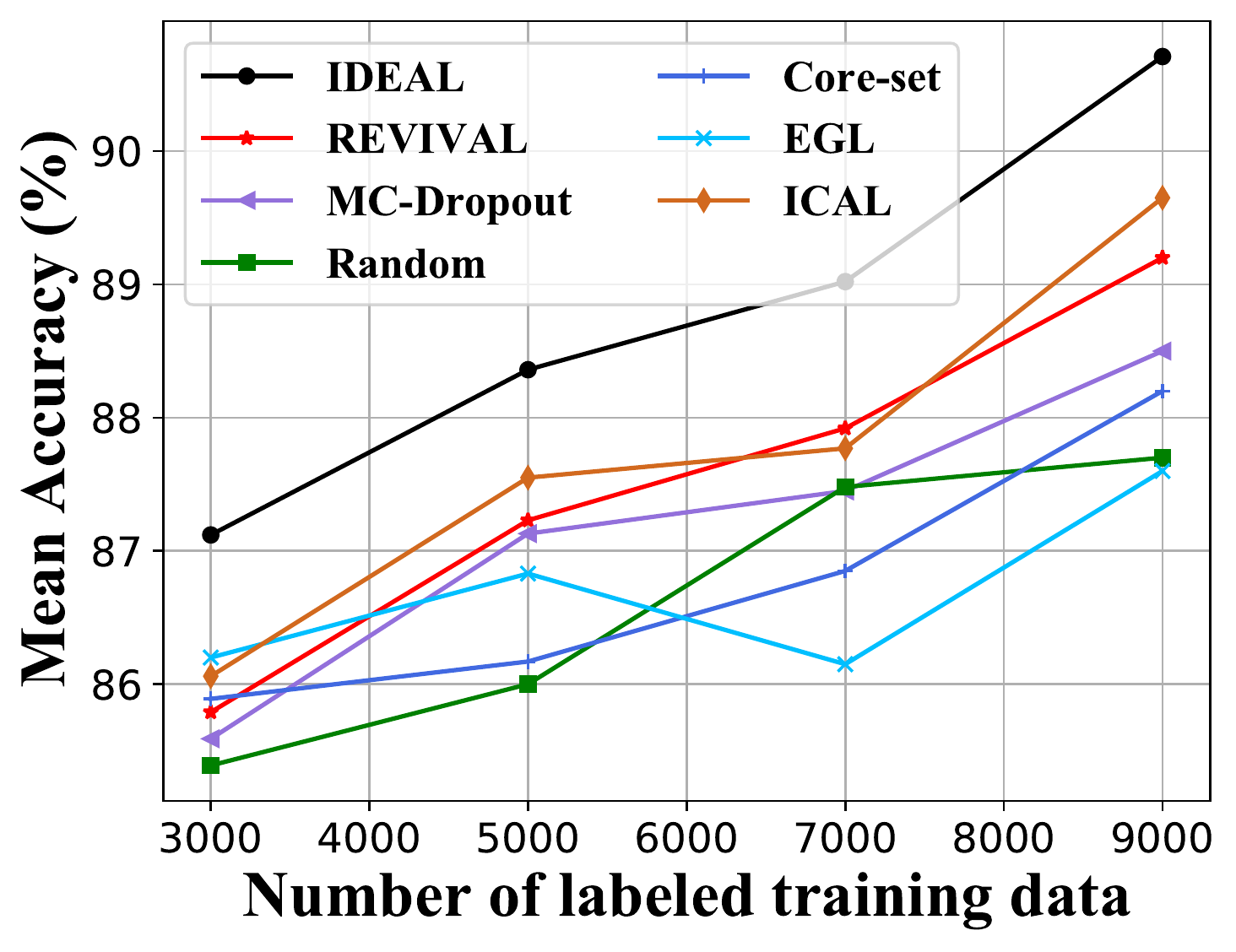}

\end{minipage}
}%
\subfigure[Bidding.]{
\begin{minipage}[t]{0.24\linewidth}
\centering
\label{semi-bid}
\includegraphics[width=1.68in]{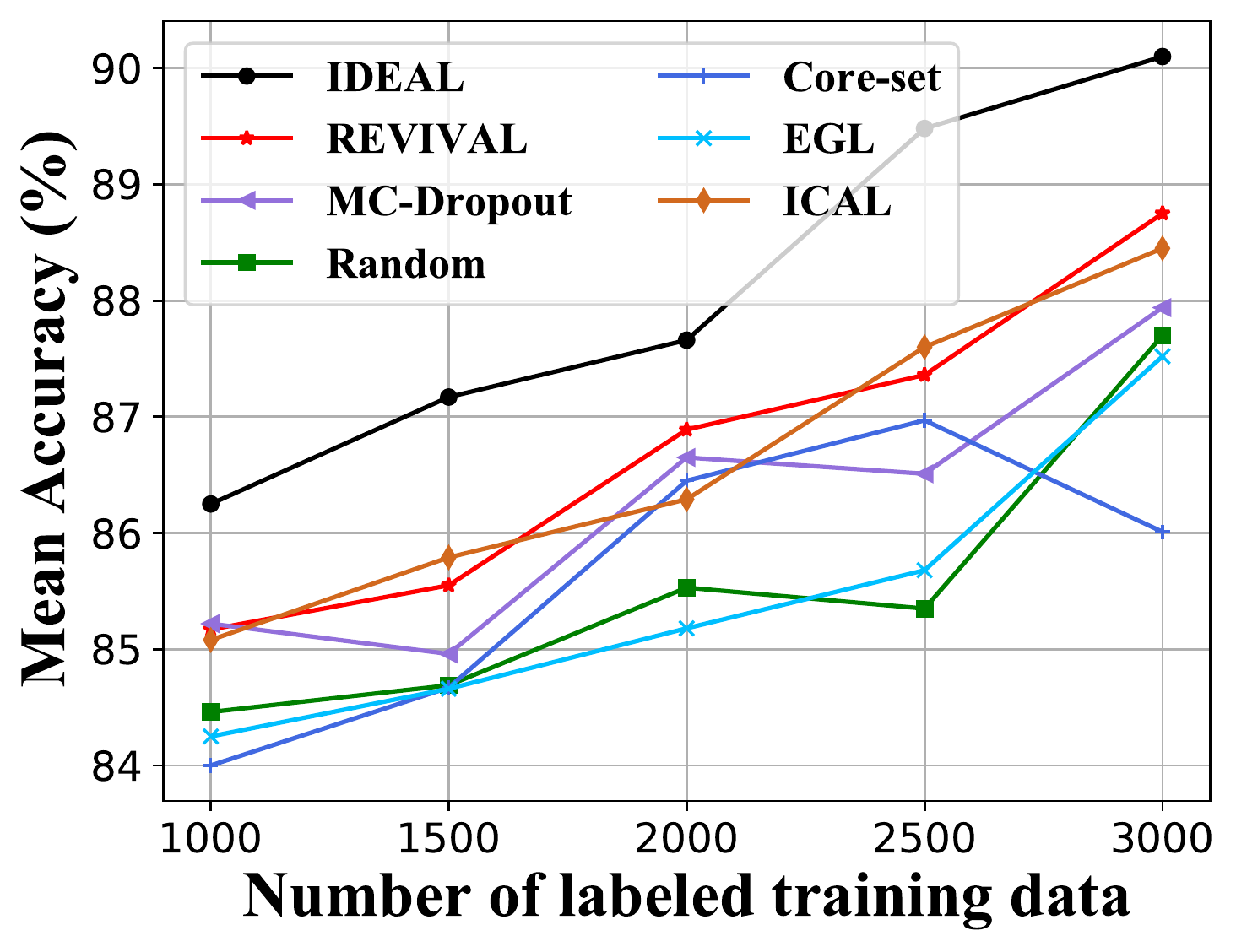}

\end{minipage}
}%
\centering
\vspace{-0.5cm}
\caption{Algorithm performance in real-world scenarios. Results in (a) and (b) are obtained under supervised learning. Results in (c) and (d) are obtained under semi-supervised learning.}
\label{real}
\vspace{-0.2cm}
\end{figure*}

\noindent $\bullet$ \textbf{Cost saving analysis of industrial application.}\quad 
To address IDEAL's potential advancement on industrial applications, we perform an in-depth performance plus cost-saving analysis on two real-world industrial text datasets. As shown in Figures~\ref{law}, ~\ref{semi-law} and Figures~\ref{bid}, ~\ref{semi-bid}, IDEAL maintains consistent superiority over other SOTAs in all experiment settings. In real scenarios, AL is used to improve the system's performance to a certain usable standard at the cost of minimal annotations. From this perspective, a more practical way to measure AL's performance is to compare the annotation cost saving (Labeling cost per label ($\$$/$L$) $\times$ Number of labels ($N$) ) demanded to reach a specific system performance. 
Table~\ref{tab:selected_data} summarizes the statistics of annotation costs:  
1) \textbf{Legal dataset}. IDEAL reaches 88.36\% accuracy with 5,000 labeled samples, reducing 44.44\% (\textbf{IDEAL \emph{vs} Random sampling, saving $\$$3640}) and 28.57\% (\textbf{{IDEAL \emph{vs} ICAL, } {saving $\$$1820}}) annotations, respectively.
2) \textbf{Bidding dataset}. IDEAL achieves 86.25\% and 87.70\% accuracy with 1,000 and 2,000 labeled samples, respectively. With the better performance, IDEAL reduces 60.0\% (\textbf{{IDEAL \emph{vs} Random sampling, saving $\$$1380}}) and 20.0\% (\textbf{IDEAL {\emph{vs} ICAL, saving $\$$460}}) of annotation cost. In sum, IDEAL is promising to reduce the annotation cost significantly for various industrial AI efforts. Notably, the analysis also shows that the data magnitude and cost-cutting scale are positively correlated. In other words, as the data required from industrial tasks grows, IDEAL can investigate unlabeled data deeply, and the annotation cost saving can rise in lockstep with the data scale. For instance, the number of unlabeled samples in a legal fact-finding dataset exceeds 10 million; IDEAL can expedite the data annotation process and save a few hundred thousand dollars in the best-case scenario.


\section{MODEL ANALYSIS}
To further validate the proposed IDEAL's effectiveness, we select the most commonly used dataset (CIFAR-100) and perform an in-depth analysis of our method on the dataset.
\vspace{-0.2cm}

\subsection{Robustness to Hyperparameters}
\begin{figure}[htbp]
\centering
\subfigure[$\mathcal{M}$.]{

\begin{minipage}[t]{0.48\linewidth}
\centering
\label{hyper_m}
\includegraphics[width=1.58in]{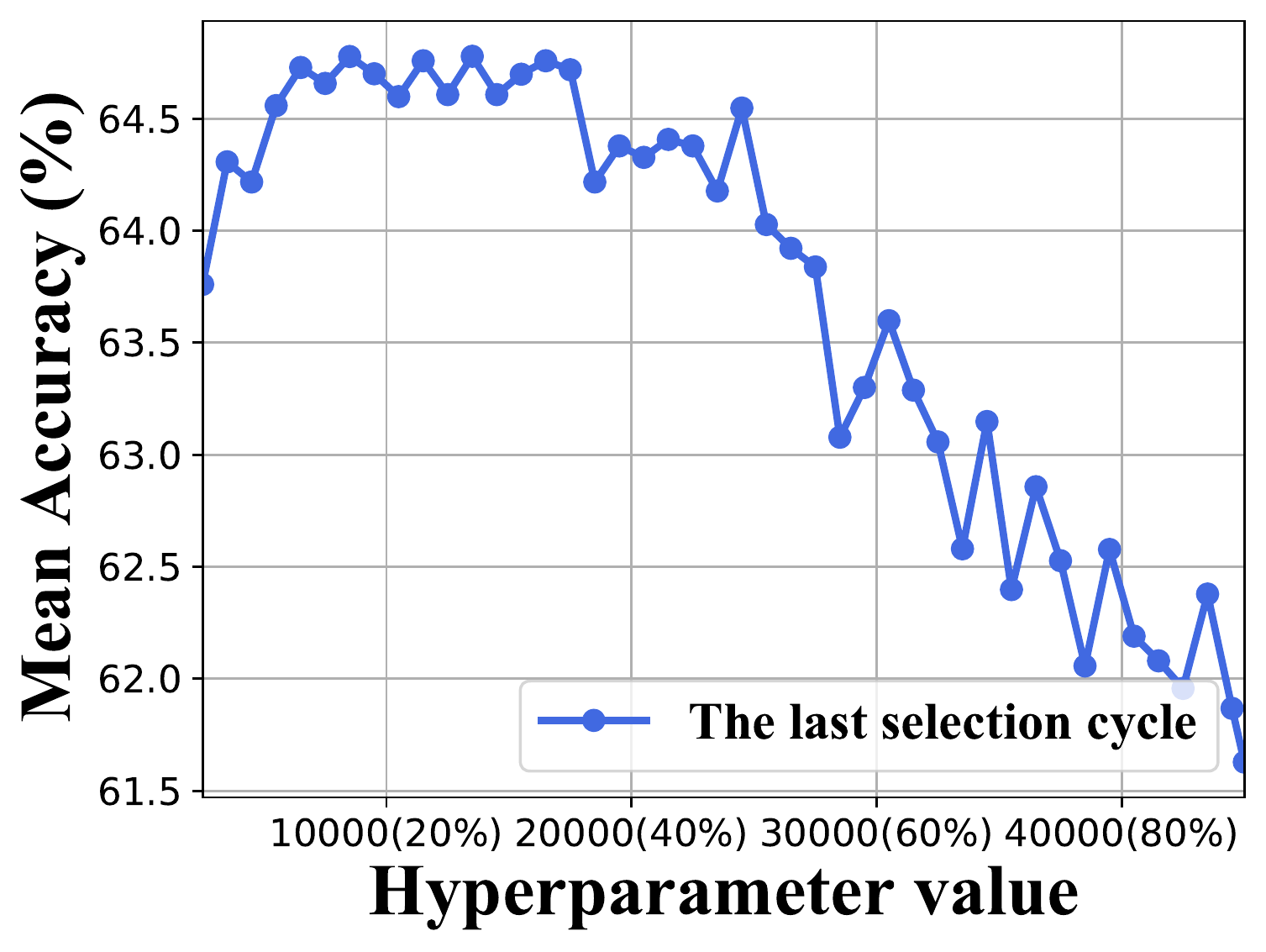}

\end{minipage}%
}%
\subfigure[$\gamma$.]{
\begin{minipage}[t]{0.48\linewidth}
\centering
\label{hyper_r}
\includegraphics[width=1.58in]{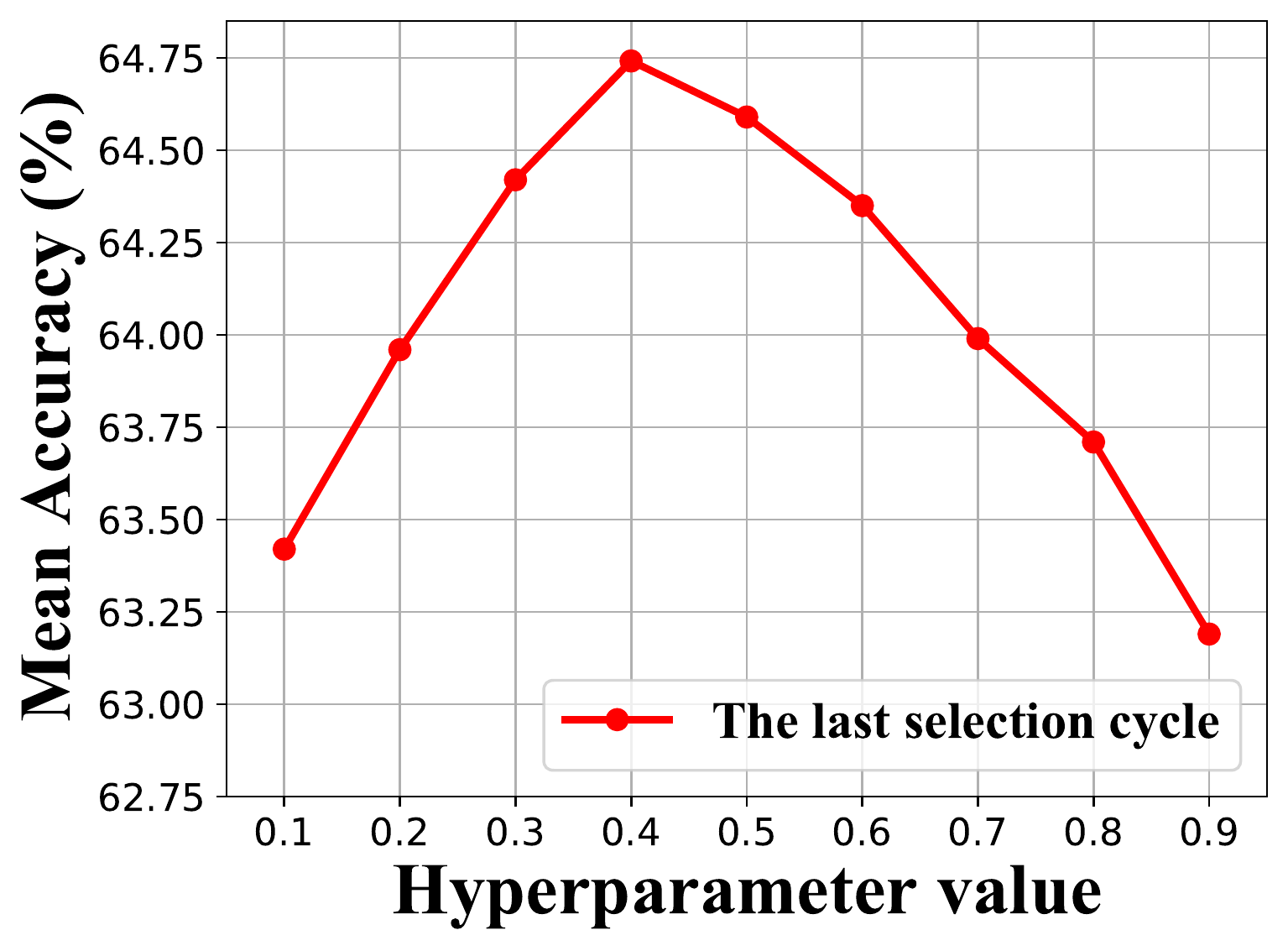}

\end{minipage}%
}%

\centering
\vspace{-0.4cm}
\caption{Performance under different hyper-parameters of $\mathcal{M}$ and $\gamma$ on CIFAR-100 dataset.}
\vspace{-0.7cm}
\end{figure}

\noindent $\bullet$ \textbf{Hyperparameter $\mathcal{M}$.}\quad 
We conduct experiments to analyze the proposed method's robustness to hyper-parameter $\mathcal{M}$ against the CIFAR-100 dataset under supervised learning. The results are shown in Figure~\ref{hyper_m}. In each selection stage, the primary selection is made by inconsistency, and then the further selection is made by entropy. Their impacts on performance vary with $\mathcal{M}$. Besides, top-$\mathcal{K}$ is not a hyperparameter as it is equal to the budget size (2500) in each cycle. 
When $\mathcal{M}=2500$ (the number of samples in primary selection equals the budget), only inconsistency affects the performance. When $\mathcal{M}$ equals the number of all unlabeled samples (equivalent to no primary selection), only entropy affects the performance. When $\mathcal{M}$ is in between, inconsistency and entropy work together. From $\mathcal{M}=6500$ to $\mathcal{M}=17500$, the impact of inconsistency and entropy on performance is relatively stable (IDEAL has strong robustness to $\mathcal{M}$). After $\mathcal{M}=17500$, the impact of inconsistency begins to decrease, and correspondingly, the effect of entropy begins to increase. We tend to choose smaller $\mathcal{M}$ to save computational overhead while ensuring accurate performance. Thus, we set $\mathcal{M}$ to 6500. We can obtain similar results for other datasets.


\noindent $\bullet$ \textbf{Hyperparameter $\gamma$.}\quad 
We conduct an exploratory analysis to systematically infer a proper weight coefficient $\gamma$ for balancing coarse-grained and fine-grained inconsistency. On the one hand, we can observe from Figure~\ref{hyper_r} that IDEAL obtains the best performance (accuracy of 64.74\%) when $\gamma$=0.4, \emph{i.e.,} the weight of coarse-grained and fine-grained inconsistency are 0.4 and 0.6, respectively. These results suggest that the fine-grained inconsistency has a higher weight to select informative AL samples. This phenomenon is reasonable because the fine-grained perturbation (pixel-level or embedding-level) can guide the model to deeply explore continuous local distribution non-smooth deeply and select the non-smooth samples. On the other hand, using the coarse-grained or fine-grained inconsistency as the major guidance ($\gamma$=0.1 or 0.9) to choose AL samples will yield relatively unsatisfactory results. The findings indicate the importance of balance between coarse-grained or fine-grained inconsistency. In other words, the two granularities of inconsistency in IDEAL can work together to pick the informative unlabeled samples as AL annotation candidates for superior SSL in the following learning cycle.

\subsection{Ablation Study}

We present ablation studies to evaluate the contribution of critical modules of IDEAL on the CIFAR-100 dataset in Table~\ref{ablation}.
For convenience, we use Ranker and Re-ranker to represent virtual inconsistency ranker and density-aware uncertainty re-ranker respectively. Without the ranker, the model accuracy yields a significant drop ($72.23\%$$\rightarrow$$70.19\%$). Diving into the ranker, coarse-grained inconsistency estimation helps SSL exclude unstable samples (not robust to augmentation methods), while fine-grained inconsistency estimation helps SSL exclude unsmoothed samples (sensitive to adversarial perturbation). Besides, the continuous exploration of local distribution is more beneficial than limited and discrete data augmentations ($71.38\%$$\rightarrow$$71.09\%$). Moreover, the re-ranker can further constrain the entropy of initially selected samples and take samples' feature distribution into account. Based on our proposed density-aware uncertainty, the re-ranker further boosts model performance from $71.68\%$ to $72.23\%$.

\begin{table}[h]
\small
\vspace{-0.3cm}
\caption{Ablation study under semi-supervised learning over CIFAR-100. Symbol - indicates that IDEAL excludes the corresponding module.}

  {
    \begin{tabular}{ l c c c c c c c c}
     \toprule[1.5pt]
	 \multicolumn{5}{c|}{\multirow{2}{*}{\textbf{Methods}}}&\multicolumn{4}{c}{\textbf{Number of labeled samples}} \\
   \cmidrule(l){6-9}
     \multicolumn{5}{c|}{} &\multicolumn{2}{c}{7,500} &\multicolumn{2}{c}{ 12,500 } \\
    \toprule[1pt]
%
%
%
%
%
%
%
	\multicolumn{5}{l|}{\textbf{IDEAL}} &\multicolumn{2}{c}{\textbf{68.46$\pm$0.17}} &\multicolumn{2}{c}{\textbf{72.23$\pm$0.20}}\\
	\toprule[1pt]
	\multicolumn{5}{l|}{- density-aware module }	&\multicolumn{2}{c}{68.24$\pm$0.21} &\multicolumn{2}{c}{71.93$\pm$0.19}\\
	\multicolumn{5}{l|}{- Re-ranker}	&\multicolumn{2}{c}{68.03$\pm$0.18} &\multicolumn{2}{c}{71.68$\pm$0.18}\\
	\multicolumn{5}{l|}{- coarse-grained inconsistency}   &\multicolumn{2}{c}{67.71$\pm$0.18} &\multicolumn{2}{c}{71.38$\pm$0.19}\\
	\multicolumn{5}{l|}{- fine-grained inconsistency}	&\multicolumn{2}{c}{67.34$\pm$0.20} &\multicolumn{2}{c}{71.09$\pm$0.19}\\
	\multicolumn{5}{l|}{- Ranker}   &\multicolumn{2}{c}{66.93$\pm$0.19} &\multicolumn{2}{c}{70.19$\pm$0.17}\\
	
	\multicolumn{5}{l|}{Random}  &\multicolumn{2}{c}{66.61$\pm$0.22} &\multicolumn{2}{c}{69.82$\pm$0.22}\\

    \toprule[1.5pt]
    \end{tabular}}
    \label{ablation}
\vspace{-0.35cm}
\end{table}

\section{conclusion}
Annotated data is the backbone of vast ML algorithms, and low-resource learning, w/wo human engagement, becomes a vital task. In this study, we propose a novel inconsistency-based virtual adversarial active learning (IDEAL) framework where SSL and AL formed a unique closed-loop structure for reciprocal enhancement. IDEAL can locate optimal unlabeled samples for human annotation based on innovative coarse-grained and fine-grained inconsistency estimation and density-aware uncertainty. To validate the effectiveness and practical advantages of IDEAL, we conducted experiments over several benchmarks and real-world datasets. Experimental results across text and image domains witness that IDEAL could make significant improvements over existing SOTA methods. In the future, we will investigate more sophisticated models to enable enhanced collaborations between human and algorithm, \emph{e.g.,} using explainable SSL to maximize human annotators' contribution.

\begin{acks}
This work has beed supported in part by National Key Research and Development Program of China 
(2018AAA0101900), Zhejiang NSF (LR21F020004), Alibaba-Zhejiang University Joint Research Institute of Frontier Technologies, Alibaba Group through Alibaba Research Intern Program, Key Research and Development Program of Zhejiang Province, China (No.2021C01013), Key Research and Development Plan of Zhejiang Province (Grant No.2021C03140), Chinese Knowledge Center of Engineering Science and Technology (CKCEST).

The author Guo would also like to thank his girlfriend, Miss. Earth  (Xiaolin Li), for the considerable support in plotting figures.
\end{acks}

\bibliographystyle{ACM-Reference-Format}
\bibliography{sample-base}
%
%
%
%
%
%
%
%
%
\clearpage
\appendix

\section{APPENDIX}
\subsection{Algorithm Details}\label{alg_sec}

\begin{algorithm}[h]
	\renewcommand{\algorithmicrequire}{\textbf{Input:}}
	\renewcommand{\algorithmicensure}{\textbf{Output:}}
	\caption{\textbf{I}nconsistency-based virtual a\textbf{D}v\textbf{E}rsarial \textbf{A}ctive \textbf{L}earning (IDEAL)}
	\begin{algorithmic}[1]
		\Require Labeled pool $\mathcal{D}^l$, Unlabeled pool $\mathcal{D}^u$, Task model's parameters $p(y\arrowvert x,\theta)$, Initial candidates set's size $\mathcal{M}$, Budget $\mathcal{K}$, Selection epochs T
		\Ensure Task model trained with $\mathcal{D}^l$ and $\mathcal{D}^u$ \\
		{\bf Initialize}
		$\mathcal{D}^l$ by randomly sampling
		\For{t = 0 to T-1}
		\State $\textbf{X}^u \leftarrow x^u$ 
		\algorithmiccomment{\textcolor{black}{Coarse-grained augmentation}}
		\State $\Tilde{\textbf{X}}^u \leftarrow \textbf{X}^u$ \algorithmiccomment{\textcolor{black}{Fine-grained augmentation}}
		\State $(\Tilde{\textbf{Y}}^u, \Tilde{y}^u ) \leftarrow p(\Tilde{\textbf{X}}^u\cup x^u,\theta)$ \algorithmiccomment{\textcolor{black}{Infer labels}}
		\State $\Tilde{y}^a \leftarrow \textit{ave}(\Tilde{\textbf{Y}}^u, \Tilde{y}^u )$ \algorithmiccomment{\textcolor{black}{Average labels}}
		\State $(y^{'},x^{'}) \leftarrow  \textit{mix}((x^l,y^l),(x^u,\Tilde{y}^a),(\Tilde{\textbf{X}}^u,\Tilde{y}^a))$ \algorithmiccomment{\textcolor{black}{Mix up}}
		\State $p(y\arrowvert x,\theta)\leftarrow (y^{'},x^{'})$ \algorithmiccomment{\textcolor{black}{Train the task model}}
		\State $\bar{\textbf{X}}^u \leftarrow x^u$ \algorithmiccomment{\textcolor{black}{Coarse-grained augmentation}}
		\State $(\bar{\textbf{Y}}^u,\bar{y}^u) \leftarrow p(\bar{\textbf{X}}^u\cup x^u,\theta)$ \algorithmiccomment{\textcolor{black}{Infer labels}}
		\State ${\textbf{R}}_{adv} \leftarrow (\bar{\textbf{X}}^u,\bar{\textbf{Y}}^u)$\algorithmiccomment{\textcolor{black}{Compute perturbation}}
		\State $\hat{\textbf{X}}^u \leftarrow \bar{\textbf{X}}^u+\textbf{R}_{adv}$\algorithmiccomment{\textcolor{black}{Fine-grained augmentation}}
		\State $\hat{\textbf{Y}}^u \leftarrow p(\hat{\textbf{X}}^u,\theta)$\algorithmiccomment{\textcolor{black}{Infer labels}}
		\State ${In}(x^u) \leftarrow (KL(\bar{\textbf{Y}}^u,\hat{\textbf{Y}}^u), Var(\bar{\textbf{Y}}^u\cup \bar{y}^u)) $ 
			\algorithmiccomment{\textcolor{black}{Inconsistency}}
		\State 	Select top $\mathcal{M}$ candidates with the largest ${In}(x^u)$
		\State 	Compute ${En}(x^m_i)$ for each $x^m_i$ in $\mathcal{M}$  
			\algorithmiccomment{\textcolor{black}{weighted entropy}}
		\State 	Select top $\mathcal{K}$ samples as finally selected set $\mathcal{D}^s$ with the largest density-aware entropy for human annotation 
		\State 	Update $\mathcal{D}^l=\mathcal{D}^l\cup \mathcal{D}^s$, $\mathcal{D}^u=\mathcal{D}^u/ \mathcal{D}^s$ 
		\EndFor
		\State  $\textbf{end for}$ \\
		\Return Task model trained with $\mathcal{D}^l$ and $\mathcal{D}^u$
	\end{algorithmic}
	
\label{alg}
\end{algorithm}

\subsection{Additional Details on the Dataset}\label{dataset}
Table~\ref{statistics} summarizes the statistics of datasets used in experiments. The legal dataset consists of the fact-finding portion of the public adjudication documents. The case type of the dataset is private lending disputes (PLD). We collect the bidding dataset from public bid notifications to classify enterprises' purchase and sale documents. We leverage the bidding dataset to help customers search for desired bidding information. 

\begin{table}[h]
\caption{Datasets statistics.}
  \centering
  {
    \begin{tabular}{ c c c c c c}
    \hline
     \multicolumn{1}{c}{Datasets} &\multicolumn{1}{c}{Type} &\multicolumn{1}{c}{Classes} &\multicolumn{1}{c}{ Train}&\multicolumn{1}{c}{ Val }&\multicolumn{1}{c}{ Test }\\

 	\hline
 	CIFAR-10 & Image & 10& 45,000& 5,000& 10,000\\
 	CIFAR-100 & Image & 100& 45,000&  5,000& 10,000\\
 	AG News & Text & 4& 112,000& 8,000& 7,600\\
 	IMDB   & Text & 2& 20,000& 5,000& 25,000\\
 	Legal  & Text & 12& 27,432& 1,523& 1,524\\
 	Bidding  & Text & 22& 23,000& 2,000& 2,514\\
    \hline
    \end{tabular}
    }
    \label{statistics}
\end{table}

\subsection{Additional Details on the Baselines}\label{baseline}
The baselines used in our experiments can be summarized as follows: 

\begin{itemize}
	\item
	EGL~\cite{zhang2017active} selects samples with the largest expected gradient change, as they are expected to impose a large impact on the model. The expectation is computed over the posterior distribution of labels for the sample according to the trained model.
	\item
	Core-set~\cite{sener2017active} is a distribution-based sampling algorithm that selects a subset to cover the whole set’s distribution. The relation between the subset and the whole set is shown in~\cite{sener2017active} intuitively from the perspective of geometry, and the NP-hard problem of subset selection can be solved efficiently by a greedy algorithm.
	\item MC-dropout~\cite{siddhant2018deep} selects samples with the largest uncertainty that is calculated by Monte Carlo Dropout on multiple inference cycles. It uses the max-entropy acquisition function.
	\item Random~\cite{figueroa2012active} sampling randomly selects samples and often serves as the lower bound of active learning algorithms.
	\item VAAL~\cite{sinha2019variational} learns representations of both labeled and unlabeled data in latent space. Afterwards, the VAE-GAN module uses extracted representations to estimate the label state information and pick the most informative samples through a min-max adversarial training process.  
	
	\item SRAAL~\cite{zhang2020state} deeply explores the label state information and maps discrete label states into a continuous variable with a state relabeling method, compared to VAAL.

	\item ICAL~\cite{gao2020consistency} chooses samples with the highest inconsistency of predictions over a set of data augmentations.
	\item LLAL~\cite{yoo2019learning} annotates samples with the largest training loss.
	\item REVIVAL~\cite{guo2021semi} utilizes a graph to boost AL with an effective prior by inferring and passing samples' relationships through the graph. It annotates samples near the boundary of clusters in the graph, which cloud refine the graph faster as a better prior for SSL.
\end{itemize}

%
%

\subsection{Computational Costs Analysis}
In this subsection, we conduct experiments on computational cost for IDEAL and its contemporary hierarchical sampling method REVIVAL~\cite{guo2021semi}. The experiments are conducted on a Linux server equipped with an Intel(R) Xeon(R) CPU E5-2690 v4 @ 2.60GHz, 512GB RAM and two NVIDIA Titan V GPU. We use the Yahoo! Answers dataset (1,400,000 training data)~\cite{chang2008importance} to exhaustedly test the computational cost of methods by changing the size of the unlabeled pool. We fine-tune the hyper-parameters of these two methods to reach their peak performance for a fair comparison. The results (average over five experiment trials) of actual running time and additional memory cost are shown in Table~\ref{cost}. We can observe that REVIVAL consumes abundant computing resources, including actual running time and additional memory space, to build a KNN graph. At the same time, IDEAL is able to deal with a huge unlabeled pool at a higher speed and memory-efficient. Thus, IDEAL is more advantageous when facing large-scale, high-dimensional datasets and deployed in practical industry scenarios.   

\begin{table*}[hb]
\caption{Computational costs comparison on the Yahoo! Answers dataset.}

  {
    \begin{tabular}{ c c c c c c c c c c}
     \toprule[1.5pt]
	 \multicolumn{2}{c|}{\multirow{2}{*}{\textbf{Data size}}}&\multicolumn{4}{c|}{\textbf{IDEAL}} &\multicolumn{4}{c}{\textbf{REVIVAL}}\\
   \cmidrule(l){3-6} \cmidrule(l){7-10}
     \multicolumn{2}{c|}{} &\multicolumn{2}{c|}{Running time (s)} &\multicolumn{2}{c|}{Memory cost (MB)} &\multicolumn{2}{c|}{Running time (s)} &\multicolumn{2}{c}{Memory cost (MB)} \\
    \toprule[1pt]
    \multicolumn{2}{c|}{20,000}&\multicolumn{2}{c}{14.1}&\multicolumn{2}{c|}{ 35 }&\multicolumn{2}{c}{191.2}&\multicolumn{2}{c}{1376}\\
    \multicolumn{2}{c|}{60,000}&\multicolumn{2}{c}{41.1}&\multicolumn{2}{c|}{ 91 }&\multicolumn{2}{c}{499.3}&\multicolumn{2}{c}{3782}\\
    \multicolumn{2}{c|}{100,000}&\multicolumn{2}{c}{65.2}&\multicolumn{2}{c|}{ 138 }&\multicolumn{2}{c}{923.0}&\multicolumn{2}{c}{6022}\\
    \multicolumn{2}{c|}{150,000}&\multicolumn{2}{c}{101.1}&\multicolumn{2}{c|}{ 204 }&\multicolumn{2}{c}{1722.9}&\multicolumn{2}{c}{8965}\\
    \multicolumn{2}{c|}{200,000}&\multicolumn{2}{c}{131.2}&\multicolumn{2}{c|}{ 283 }&\multicolumn{2}{c}{1538.5}&\multicolumn{2}{c}{11652}\\
    \multicolumn{2}{c|}{400,000}&\multicolumn{2}{c}{267.3}&\multicolumn{2}{c|}{ 574 }&\multicolumn{2}{c}{3175.1}&\multicolumn{2}{c}{22806}\\
    \multicolumn{2}{c|}{600,000}&\multicolumn{2}{c}{404.2}&\multicolumn{2}{c|}{ 862 }&\multicolumn{2}{c}{4913.0}&\multicolumn{2}{c}{34576}\\
    \multicolumn{2}{c|}{800,000}&\multicolumn{2}{c}{479.3}&\multicolumn{2}{c|}{ 1069 }&\multicolumn{2}{c}{6374.5}&\multicolumn{2}{c}{45927}\\
    \multicolumn{2}{c|}{900,000}&\multicolumn{2}{c}{545.7}&\multicolumn{2}{c|}{1175}&\multicolumn{2}{c}{7296.8}&\multicolumn{2}{c}{51642}\\
%
%
%
%
%
%
%
%
%
    \toprule[1.5pt]
    \end{tabular}}
    \label{cost}
\end{table*}

\end{document}